\documentclass[sigconf]{acmart}
\usepackage[ruled, vlined, linesnumbered]{algorithm2e}
\usepackage{microtype}
\usepackage{bm}
\usepackage{graphicx}
\usepackage{amsmath}
\usepackage{multirow}
\usepackage{cleveref}
\usepackage{xspace}
\usepackage{xcolor}
\usepackage{enumitem}
\usepackage{placeins}
\usepackage{soul}
\usepackage{bbm}
\usepackage{booktabs}
\usepackage{hyperref}
\usepackage{subcaption}
\usepackage[utf8]{inputenc}
\usepackage{pgfplots}
\DeclareUnicodeCharacter{2212}{−}
\usepgfplotslibrary{groupplots,dateplot}
\usetikzlibrary{patterns,shapes.arrows}
\pgfplotsset{compat=newest}

\newcommand{\ours}{{CANMD}\xspace}

\AtBeginDocument{%
  \providecommand\BibTeX{{%
    \normalfont B\kern-0.5em{\scshape i\kern-0.25em b}\kern-0.8em\TeX}}}

\copyrightyear{2022} 
\acmYear{2022} 
\setcopyright{acmlicensed}\acmConference[CIKM '22]{Proceedings of the 31st ACM International Conference on Information and Knowledge Management}{October 17--21, 2022}{Atlanta, GA, USA}
\acmBooktitle{Proceedings of the 31st ACM International Conference on Information and Knowledge Management (CIKM '22), October 17--21, 2022, Atlanta, GA, USA}
\acmPrice{15.00}
\acmDOI{10.1145/3511808.3557263}
\acmISBN{978-1-4503-9236-5/22/10}

\begin{document}

\title{Contrastive Domain Adaptation for Early Misinformation Detection: A Case Study on COVID-19
}



\author{Zhenrui Yue}
\affiliation{%
  \institution{University of Illinois Urbana-Champaign}
  \country{USA}}
\email{zhenrui3@illinois.edu}

\author{Huimin Zeng}
\affiliation{%
  \institution{University of Illinois Urbana-Champaign}
  \country{USA}}
\email{huiminz3@illinois.edu}

\author{Ziyi Kou}
\affiliation{%
  \institution{University of Illinois Urbana-Champaign}
  \country{USA}}
\email{ziyikou2@illinois.edu}

\author{Lanyu Shang}
\affiliation{%
  \institution{University of Illinois Urbana-Champaign}
  \country{USA}}
\email{lshang3@illinois.edu}

\author{Dong Wang}
\affiliation{%
  \institution{University of Illinois Urbana-Champaign}
  \country{USA}}
\email{dwang24@illinois.edu}

\renewcommand{\shortauthors}{Zhenrui Yue et al.}

\begin{abstract}
Despite recent progress in improving the performance of misinformation detection systems, classifying misinformation in an unseen domain remains an elusive challenge. To address this issue, a common approach is to introduce a domain critic and encourage domain-invariant input features. However, early misinformation often demonstrates both conditional and label shifts against existing misinformation data (e.g., class imbalance in COVID-19 datasets), rendering such methods less effective for detecting early misinformation. In this paper, we propose contrastive adaptation network for early misinformation detection (\ours). Specifically, we leverage pseudo labeling to generate high-confidence target examples for joint training with source data. We additionally design a label correction component to estimate and correct the label shifts (i.e., class priors) between the source and target domains. Moreover, a contrastive adaptation loss is integrated in the objective function to reduce the intra-class discrepancy and enlarge the inter-class discrepancy. As such, the adapted model learns corrected class priors and an invariant conditional distribution across both domains for improved estimation of the target data distribution. To demonstrate the effectiveness of the proposed \ours, we study the case of COVID-19 early misinformation detection and perform extensive experiments using multiple real-world datasets. The results suggest that \ours can effectively adapt misinformation detection systems to the unseen COVID-19 target domain with significant improvements compared to the state-of-the-art baselines.
\end{abstract}

\ccsdesc[500]{Information systems~Data mining}
\ccsdesc[500]{Computing methodologies~Natural language processing}

\keywords{misinformation detection, domain adaptation}


\maketitle


\section{Introduction}
\label{sec:intro}

Recent progress in developing natural language processing (NLP) models leads to significant improvements in text classification performance~\cite{devlin-etal-2019-bert, liu2019roberta}. Nevertheless, misinformation detection remains one of the most elusive challenges in text classification, especially when the model is trained on a source domain but deployed to classify misinformation in a different target domain~\cite{du-etal-2020-adversarial, zou-etal-2021-unsupervised, zeng2022attacking}. A recent example can be found in the early spreading stage of the COVID-19 pandemic, where misinformation detection systems often fail to distinguish valuable information from large amounts of rumors and misleading posts on social media platforms, resulting in potential threats to public health and interest~\cite{li2021multi}.


\begin{figure}[h]
\centering
\includegraphics[width=0.9\linewidth]{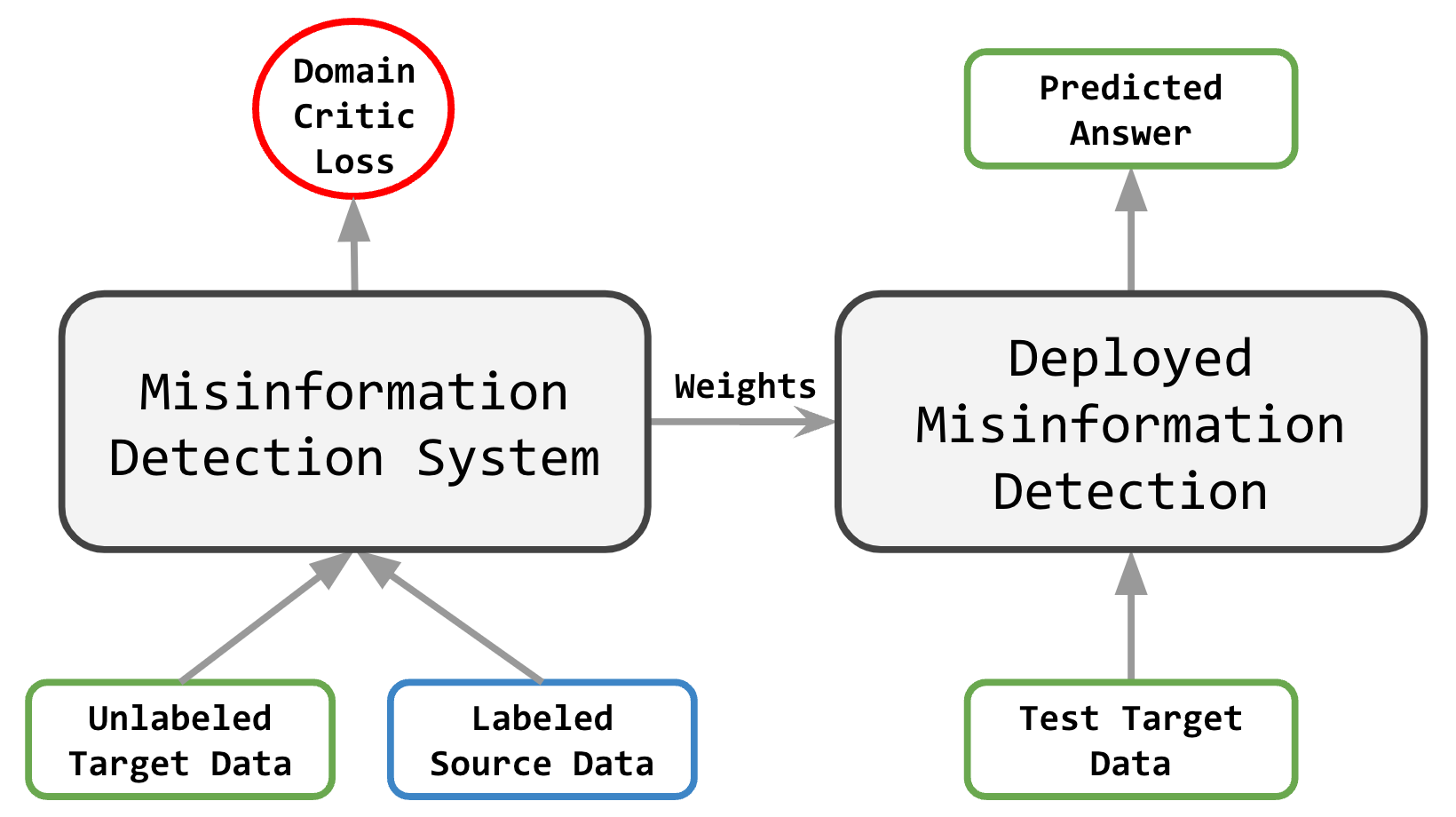}
\caption{Labeled source data and unlabeled target data are accessible for domain adaptation. Upon deployment, the model predicts misinformation in the target domain.} 
\label{fig:intro}
\vspace{-5pt}
\end{figure}

In this paper, we study the domain adaptation problem in early misinformation detection using COVID-19 data. Widespread COVID misinformation poses a major threat to the online ecosystem and potentially dangers the public health. For example, \citet{roozenbeek2020susceptibility} demonstrate strong correlation between COVID-19 misinformation and noncompliance of health guidance as well as reduced likelihood in receiving vaccines. To identify potential misinformation on social media platforms, one possible solution is to train supervised models via external knowledge or crowdsourcing~\cite{pan2018content, shu2019defend, muller2020covid, roitero2020covid, hossain-etal-2020-covidlies, medina-serrano-etal-2020-nlp, rashid2021covidsens, kou2021fakesens, shang2021multimodal, kou2022hc, shang2022duo, shang2022privacy, kou2022crowd}. Yet such learning approaches require large amounts of annotations or extensive computational resources, making existing methods ineffective for early misinformation detection in COVID-19, where label-rich misinformation data and domain knowledge are inaccessible for training~\cite{li2021multi}.

As a solution, domain adaptation methods can be used to adapt an existing misinformation detection system trained on the label-rich source domain to a label-scarce target domain~\cite{du-etal-2020-adversarial, zou-etal-2021-unsupervised, li2021multi}. \Cref{fig:intro} gives an overview of the domain adaptation in misinformation detection. One common approach is to estimate the domain discrepancy (e.g., via an additional discriminator) and impose penalty when source and target features are distinguishable. As such, the model learns domain-invariant features, and thereby improving the classification performance in the target domain~\cite{ganin2016domain, tzeng2017adversarial, kang2019contrastive}. However, early misinformation of COVID-19 often demonstrates both conditional shift (i.e., $p(\bm{x}|y) \neq q(\bm{x}|y)$) and label shift (i.e., $p(y) \neq q(y)$) compared to existing misinformation datasets. Consequently, such large discrepancies often limit the adaptation performance or even lead to negative transfer using the aforementioned methods (see \Cref{sec:experiment}), rendering the existing methods less effective for early misinformation detection in COVID-19.

To improve the adaptation performance under large domain discrepancies, we propose contrastive adaptation network for early misinformation detection (\ours). Specifically, we leverage a pretrained misinformation detection system to generate labeled target examples via pseudo labeling. To correct the label shift between pseudo labels and the target output distributions, we design a learnable rescaling component to estimate and correct the output probabilities of the pseudo labels. Then, joint training is performed by sampling source and target examples using a class-aware sampling strategy to resemble the target data distribution. Here, we propose a contrastive adaptation loss to reduce the discrepancy for intra-class data examples and enlarge the discrepancy for inter-class data examples. As such, the adapted misinformation detection system learns an invariant conditional distribution. Combined with the corrected class priors, we improve the adaptation performance by better approximating the joint distribution in the target domain. To demonstrate the effectiveness of the proposed \ours, we study the case of COVID-19 early misinformation detection and perform extensive experiments with real-world datasets. In particular, we adopt five source misinformation datasets published before the COVID outbreak. We additionally adopt three COVID misinformation datasets from 2020 to 2022 as target datasets (i.e., CoAID, Constraint and ANTiVax~\cite{cui2020coaid, patwa2021fighting, hayawi2022anti}) to evaluate the proposed \ours. Our results suggest that \ours effectively adapts misinformation detection systems to the COVID-19 target domain and consistently outperforms state-of-the-art baselines with an average improvement of 11.5\% in balanced accuracy (BA).

We summarize the main contributions of our works\footnote{Our implementation is publicly available at https://github.com/Yueeeeeeee/CANMD.}:
\begin{enumerate}
\item We propose a learnable rescaling component in \ours to correct the pseudo labels for improved estimation of the class priors. Additionally, we design a class-aware sampling strategy to resemble the target domain data distribution.
\item \ours learns a domain-invariant conditional distribution by reducing the intra-class discrepancy and enlarging the inter-class discrepancy via contrastive learning. Combined with the corrected priors, we improve the adaptation by approximating the joint distribution of the target data.
\item To the best of our knowledge, \ours is the first work that adopts pseudo labeling with label correction and contrastive domain adaptation for adapting misinformation systems to the unseen COVID-19 domain.
\item We demonstrate the effectiveness of \ours on multiple source and COVID-19 target datasets, where the proposed \ours consistently outperforms the state-of-the-art baselines by a significant margin.
\end{enumerate}
\section{Related Work}
\label{sec:related}

\subsection{Misinformation Detection}
Existing methods in misinformation detection can be mainly divided into three coarse categories: 
(1)~content-based misinformation detection: content-based models perform classification upon statements (i.e., claim) or multimodal input. For example, language models are used for misinformation detection based on linguistic features or structure-related properties~\cite{wang-2017-liar, karimi-tang-2019-learning, das2021heuristic}. Multimodal architectures are introduced to learn features from both textual and visual information for improved performance~\cite{jin2017multimodal, wang2018eann, khattar2019mvae};
(2)~social context-aware misinformation detection: \citet{jin2016news} leverage online interactions to evaluate the credibility of post contents. The propagation paths in the spreading stage can be used to detect fake news on social media platforms~\cite{liu2018early}. Other context like user dynamics and publishers are introduced to enhance social context-aware misinformation detection~\cite{guo2018rumor, shu2019beyond}; (3)~knowledge guided misinformation detection: \citet{vo-lee-2020-facts} and \citet{popat-etal-2018-declare} propose to search for external evidence to derive informative features for content verification. Multiple evidence pieces are used to build graphs for fine-grained fact verification~\cite{liu-etal-2020-fine}. Knowledge graphs introduce additional information to provide useful explanations for the detection results~\cite{cui2020deterrent, kou2022hc, kou2022crowd}. Nevertheless, the generalization and adaptability of such misinformation detection systems are not well studied. Therefore, we focus on domain adaptation of content-based language models for misinformation detection.

\subsection{Domain Adaptation in Computer Vision}
Domain adaptation methods are primarily studied on the image classification task~\cite{long2013transfer, tzeng2014deep, long2015learning, long2017deep, kang2019contrastive}. Existing approaches minimize the representation discrepancy between the source and target domains to encourage domain-invariant features, and thereby improving the model performance on shifted input distributions (i.e., covariate shifts)~\cite{long2013transfer, tzeng2014deep}. Early domain adaptation methods utilize an additional distance term in the training objective to minimize the distance between the input marginal distributions~\cite{tzeng2014deep, long2015learning}. \citet{long2013transfer} and \citet{long2017deep} adapt source-trained models by optimizing the distance between the joint distributions across both domains. Similarly, a feature discriminator can be introduced to perform minimax optimization, where the regularization on domain generalization can be implicitly imposed when source and target features are distinguishable by the discriminator~\cite{ganin2016domain, tzeng2017adversarial}. Class-aware and contrastive domain adaptation are proposed for fine-grained domain alignment, such methods adapt the conditional distributions by regularizing the inter-class and intra-class distances via an additional domain critic~\cite{pei2018multi, kang2019contrastive}. Recently, calibration and label shifts are studied for domain adaptation problems when the label distribution changes between the domains~\cite{guo2017calibration, lipton2018detecting, azizzadenesheli2019regularized, alexandari2020maximum}. To the best of our knowledge, domain adaptation that considers both label correction and conditional distribution adaptation is not studied in current literature. Yet label shift is often observed in real-world scenarios (e.g., misinformation detection). As such, we propose an adaptation method that corrects the label shift and optimizes the conditional distribution to improve the out-of-domain performance.

\subsection{Domain Adaptation in Text Classification}
Currently, few methods are tailored for domain adaptation in misinformation detection, therefore, we review the related work in both text classification and misinformation detection. A domain-distinguishing task is proposed to post-train models for domain awareness and improved adversarial adaptation performance~\cite{du-etal-2020-adversarial}. Energy-based adaptation is proposed via an additional antoencoder that generates source-like target features~\cite{zou-etal-2021-unsupervised}. \citet{xu2019adversarial}, \citet{zhang2020bdann} and \citet{zhou2022mdmn} utilize domain adversarial training and multi-task learning to promote domain-invariant features in multimodal misinformation detection. Pseudo labeling is used to integrate domain knowledge via a weak labeling function for multi-source domain adaptation~\cite{li2021multi}. 
However, domain adaptation is not well studied for content-based misinformation detection, let alone the adaptation for COVID-related early misinformation. We develop a domain adaptation method tailored for misinformation detection \ours. Our method significantly improves the adaptation performance in COVID misinformation detection by: (1)~correcting the label shift between the source and target data distributions; and (2)~exploiting knowledge from the source domain by minimizing the distance between the source and target conditional distributions.

\begin{figure*}[t]
\centering
\includegraphics[width=0.8\linewidth]{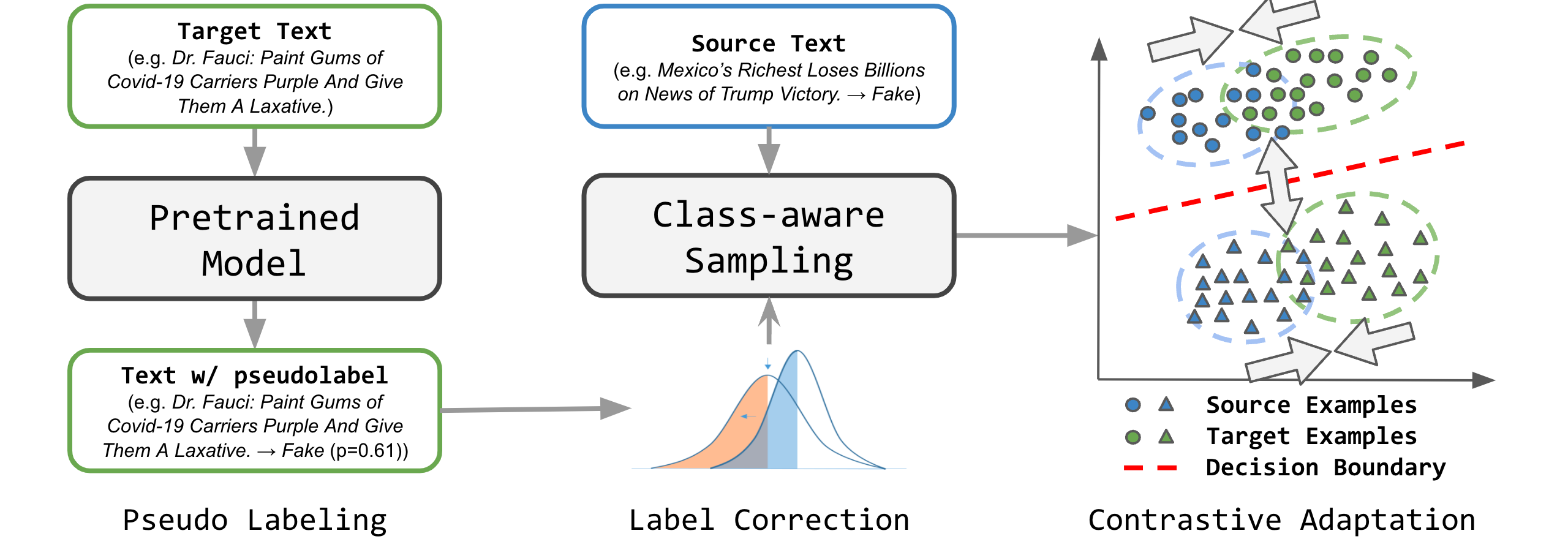}
\caption{The proposed \ours framework adapts the joint distribution in two stages: (1)~pseudo labeling and label correction, in which we approximate the target output distribution with pseudo labels generated by the pretrained model and corrected by the proposed label correction component; and (2)~contrastive adaptation, where we reduce intra-class discrepancy and enlarge inter-class discrepancy to minimize the distance between the conditional distributions from the source and target domain.} 
\label{fig:method}
\vspace{-7pt}
\end{figure*}
\section{Methodology}
\label{sec:method}

\subsection{Setup}
\textbf{Data}: Given labeled source data and unlabeled target data, our setting aims at improving the misinformation detection performance in an unseen target domain. We define misinformation detection as a binary text classification task. For this purpose, labeled source data and unlabeled target data are available, we denote the source data with $\bm{\mathcal{X}}_{s}$ and the target data with $\bm{\mathcal{X}}_{t}$. Formally, each input example is defined by a tuple consisting of an input text $\bm{x}$ and a label $y$ with $y \in \{ 0, 1 \}$. Input $\bm{x}$ is considered as misinformation (i.e., $y = 0$) if it contains primarily or entirely false information. Otherwise, the label is considered as non-misleading (i.e., $y = 1$):
\begin{itemize}
  \item \emph{Source data}: Labeled source data from $\bm{\mathcal{X}}_s$. Each example $(\bm{x}_s, y_s) \in \bm{\mathcal{X}}_s$ is defined by an input text $\bm{x}_s$ and a label $y_s$. We denote the joint distribution $(\bm{x}_s, y_s)$ in $\bm{\mathcal{X}}_s$ with $\mathcal{P}$.
  \item \emph{Target data}: Unlabeled target data from $\bm{\mathcal{X}}_t$. For target example $(\bm{x}_t, y_t) \in \bm{\mathcal{X}}_t$, we only have access to the input text $\bm{x}_t$. Ground truth label $y_t$ is not given for training. Similarly, we denote the joint distribution in $\bm{\mathcal{X}}_t$ with $\mathcal{Q}$.
\end{itemize}

\textbf{Model}: The classification model can be represented with function $\bm{f}$. $\bm{f}$ takes text $\bm{x}$ as input and yields output logits. The output probability distribution is obtained by using the softmax function $\sigma$. Ideally, the model $\bm{f}$ can correctly predict the ground truth $y$ with the maximum logit value, namely $y = \arg \max \bm{f}(\bm{x})$.

\textbf{Objective}: The objective is to adapt a classifier $\bm{f}$ trained on source distribution $\mathcal{P}$ to the target distribution $\mathcal{Q}$, such that the performance can be maximized in the target data $\bm{\mathcal{X}}_t$. In other words, we minimize the negative log likelihood (NLL) loss between model output distribution $\sigma(\bm{f}(\bm{x}_t))$ and ground truth $y_t$ for $\bm{\mathcal{X}}_t$:
\begin{equation}
    \min_{\substack{\bm{f}}} \mathbb{E}_{(\bm{x}_t, y_t) \sim \bm{\mathcal{X}}_t} \mathcal{L}_{\mathrm{nll}}(\sigma(\bm{f}(\bm{x}_t)), y_t).
\end{equation}

\subsection{Proposed Approach}
To adapt the trained classifier to the target domain, we propose contrastive adaptation network for early misinformation detection (\ours) by approximating the target joint distribution. 
Unlike previous works with the assumption of covariate shift~\cite{long2013transfer, tzeng2014deep, long2015learning, ganin2016domain}, we consider label shift (i.e., $p(y) \neq q(y)$) and conditional shift (i.e., $p(\bm{x}|y) \neq q(\bm{x}|y)$) between the source domain and target domain. Since $p(\bm{x},y) = p(\bm{x}|y)p(y)$ (Similarly, $q(\bm{x},y) = q(\bm{x}|y)q(y)$), our solution to adapt the misinformation detection model $\bm{f}$ to the target domain is two-fold: (1)~label shift estimation and correction: we generate pseudo labels for target examples in $\bm{\mathcal{X}}_t$ and rescale the labels to correct the shift in the joint distribution caused by a shift in label proportion~\cite{lipton2018detecting, azizzadenesheli2019regularized, alexandari2020maximum}. In other words, we estimate and correct $q(y)$. (2)~Conditional distribution adaptation: using the pseudo labels generated from the previous step, we minimize the distance between the conditional distributions $p(\bm{x}|y)$ and $q(\bm{x}|y)$ with contrastive domain adaptation. As we reduce the distance between both conditional distributions (i.e, $p(\bm{x}|y) \approx q(\bm{x}|y)$), we improved the estimation of $q(\bm{x}, y) = p(\bm{x}|y)q(y)$ using the corrected $q(y)$.

The reason for label shift correction lies in the rapid changing dynamics of COVID-19 misinformation. For instance, the proportion of misinformation on social media platforms changes rapidly in different stages (i.e., label shift)~\cite{cui2020coaid, patwa2021fighting, hayawi2022anti}. Additionally, conventional misinformation detection systems often fail to distinguish COVID misinformation due to large domain discrepancies, as we demonstrate in \Cref{sec:experiment}. Therefore, the proposed \ours comprises of two stages: (1)~pseudo labeling and label correction, where we perform pseudo labeling and correct the label distribution $q(y)$ in the target domain; and (2)~contrastive adaptation, in which we reduce intra-class discrepancy and enlarge inter-class discrepancy among examples from both domains. By leveraging pseudo labels from the first stage, we estimate and minimize the discrepancy between conditional distributions $p(\bm{x}|y)$ and $q(\bm{x}|y)$. 
The illustration of \ours is provided in \Cref{fig:method}, unlike existing methods, we propose to correct the label shift and adapt the conditional distributions instead of solely reducing the feature discrepancy.

\subsection{Pseudo Labeling and Label Correction}
Given access to labeled source data $\bm{\mathcal{X}}_s$, we first pretrain the misinformation classification system. In practice, we can skip this step if the model is already trained on the source data. Once the source-trained model is ready, we generate the pseudo label $\hat{y}_t$ of the unlabeled target example $\bm{x}_t$ with $\arg \max \bm{f}(\bm{x}_t)$.

However, the generated pseudo labels are noisy, which often cause deterioration of domain adaptation performance, especially when there exists a large domain gap~\cite{yue-etal-2021-contrastive}. Additionally, different label proportions between both domains (i.e., label shift) can result in biased pseudo labels and potential negative transfer results~\cite{alexandari2020maximum}. Inspired by calibration methods~\cite{guo2017calibration}, we propose a label correction component to correct the distribution shift between the pseudo labels and the target output distribution. Specifically for input $\bm{x}$, we design vector rescaling to introduce learnable parameter $\bm{w}$ and $\bm{b}$ and compute the corrected output as follows:
\begin{equation}
  \sigma(\bm{w} \odot \bm{f}(\bm{x}) + \bm{b}),
\end{equation}
where $\odot$ represents element-wise product and $\sigma$ represents the softmax function. In rare cases when the the source and target output distributions result in large bias values in $\bm{b}$, we discard the bias to avoid generating constant output probability (i.e., $\sigma(\bm{w} \odot \bm{f}(\bm{x}) + \bm{b}) \approx \sigma(\bm{b})$ when $\bm{b}$ consists of values much greater than $\bm{w} \odot \bm{f}(\bm{x})$). Thus the correction method reduces to: 
\begin{equation}
  \sigma(\bm{w} \odot \bm{f}(\bm{x})).
\end{equation}

To obtain the optimal $\bm{w}$ and $\bm{b}$, we optimize the parameter $\bm{w}$ and $\bm{b}$ in training to rescale the output probabilities and fit the distribution $q(y)$. Given pretrained misinformation detection function $\bm{f}$ and input pair $(\bm{x}_t, y_t)$, we follow~\cite{guo2017calibration} and minimize the NLL loss w.r.t. the parameters using the gradient descent algorithm, i.e.:
\begin{equation}
  \min_{\substack{\bm{w}, \bm{b}}} \mathbb{E}_{(\bm{x}_t, y_t) \sim \bm{\mathcal{X}}_t} \mathcal{L}_{\mathrm{nll}}(\sigma(\bm{w} \odot \bm{f}(\bm{x}_t) + \bm{b}), y_t),
  \label{eq:label-correction-optimization}
\end{equation}
where $\bm{b}$ is discarded in the case of constant output probability or when the optimization fails. Similar to~\cite{guo2017calibration, lipton2018detecting, alexandari2020maximum}, the validation set is used to optimize the vector rescaling parameters.

In sum, the pseudo labeling and label correction stage via the label correction component can be formulated as:
\begin{equation}
  \hat{y}_t = \arg \max \sigma(\bm{w} \odot \bm{f}(\bm{x}) + \bm{b}).
  \label{eq:label-correction}
\end{equation}
After label correction, we filter the pseudo labels to select a subset of the target examples for contrastive adaptation. The pseudo labels are filtered according to the label confidence (i.e., $\max \sigma(\bm{w} \odot \bm{f}(\bm{x}) + \bm{b})$), we preserve the target sample if the confidence value is above confidence threshold $\tau$. In the following, we denote the filtered target data with pseudo labels using $\bm{\mathcal{X}}'_t$. By introducing the label correction component, we reduce the bias from $\bm{f}$ when the source and target label distributions shift. For example, COVID misinformation in the CoAID dataset consists of less than $10\%$ false claims~\cite{cui2020coaid}. If $\bm{f}$ is pretrained on a balanced dataset, the pseudo labels would demonstrate a similar label distribution regardless of the label distribution in CoAID and result in noisy labels. \ours rescales the model output using learnable parameters to adjust the pseudo labels, followed by confidence thresholding that selects high-confidence examples from a target-similar distribution. 


\subsection{Contrastive Adaptation}
We now describe the contrastive adaptation stage of \ours. We perform mini-batch training and sample identical amounts of input pair $(\bm{x}, y)$ from both the source data $\bm{\mathcal{X}}_s$ and labeled target data $\bm{\mathcal{X}}'_t$. To approximate the target data distribution and efficiently estimate the discrepancy, we adopt a class-aware sampling strategy to guarantee the same amount of examples within each class. Specifically, a target batch is first sampled from the target data $\bm{\mathcal{X}}'_t$. By counting the examples in the target batch, we sample from the source data $\bm{\mathcal{X}}_s$ with the same number of examples from each class respectively. Consequently, we build mini-batches that comprise of the same classes and identical number of examples in both domains. We also resemble the target data distribution and incorporate the source domain knowledge during adaptation~\cite{kang2019contrastive}.

To estimate the domain discrepancy,
we revisit the maximum mean discrepancy (MMD) distance. MMD estimates the distance between two distributions using samples drawn from them~\cite{gretton2012kernel}. Given $\mathcal{P}$ and $\mathcal{Q}$, MMD distance $\mathcal{D}$ is defined as:
\begin{equation}
  \mathcal{D} = \sup_{f \in \mathcal{H}} \big( \mathbb{E}_{x \sim \mathcal{P}} [f(x)] - \mathbb{E}_{y \sim \mathcal{Q}} [f(y)] \big),
\end{equation}
where $f$ is a function (kernel) in reproducing the kernel Hilbert space $\mathcal{H}$.
By introducing the kernel trick and empirical kernel mean embeddings~\cite{long2015learning, yue-etal-2021-contrastive}, we further simplify the estimation of the squared MMD distance between $\bm{\mathcal{X}}_s$ and $\bm{\mathcal{X}}'_t$ as follows:
\begin{equation}
  \begin{aligned}
    \mathcal{D}^{\mathrm{MMD}} &= \frac{1}{|\bm{\mathcal{X}}_s||\bm{\mathcal{X}}_s|} \sum_{i=1}^{|\bm{\mathcal{X}}_s|} \sum_{j=1}^{|\bm{\mathcal{X}}_s|} k(\phi(\bm{x}_{s}^{(i)}), \phi(\bm{x}_{s}^{(j)})) \\
    &+ \frac{1}{|\bm{\mathcal{X}}'_t||\bm{\mathcal{X}}'_t|} \sum_{i=1}^{|\bm{\mathcal{X}}'_t|} \sum_{j=1}^{|\bm{\mathcal{X}}'_t|} k(\phi(\bm{x}_{t}'^{(i)}), \phi(\bm{x}_{t}'^{(j)})) \\
    &- \frac{2}{|\bm{\mathcal{X}}_s||\bm{\mathcal{X}}'_t|} \sum_{i=1}^{|\bm{\mathcal{X}}_s|} \sum_{j=1}^{|\bm{\mathcal{X}}'_t|} k(\phi(\bm{x}_{s}^{(i)}), \phi(\bm{x}_{t}'^{(j)})),
  \label{eq:contrastive_loss}
  \end{aligned}
\end{equation}
where we adopt the output from the \texttt{[CLS]} position in the transformer model as $\phi$. Here, the expectation is simplified by using mean embeddings of the drawn samples, $k$ refers to the Gaussian kernel, i.e., $k(\bm{x}_{i}, \bm{x}_{j}) = \mathrm{exp}(- \frac{\| \bm{x}_{i} - \bm{x}_{j} \|^{2}}{\gamma})$. The estimated MMD distance is used to compute a contrastive adaptation loss.
In particular, we leverage the \Cref{eq:contrastive_loss} to regularize the distances among examples within the same class and examples from different classes.
For this purpose, we define the class-aware MMD as:
\begin{equation}
  \begin{aligned}
    \mathcal{D}_{\mathrm{c_1c_2}}^{\mathrm{MMD}} &= \sum_{i=1}^{|\bm{\mathcal{X}}_s|} \sum_{j=1}^{|\bm{\mathcal{X}}_s|} \frac{\mathbbm{1}_{c_1c_2}(y_s^{(i)}, y_s^{(j)})k(\phi(\bm{x}_s^{(i)}), \phi(\bm{x}_s^{(j)}))}{\sum_{l=1}^{|\bm{\mathcal{X}}_s|} \sum_{m=1}^{|\bm{\mathcal{X}}_s|} \mathbbm{1}_{c_1c_2}(y_s^{(l)}, y_s^{(m)})} \\
    &+ \sum_{i=1}^{|\bm{\mathcal{X}}'_t|} \sum_{j=1}^{|\bm{\mathcal{X}}'_t|} \frac{\mathbbm{1}_{c_1c_2}(y_t^{\prime(i)}, y_t^{\prime(j)})k(\phi(\bm{x}_t^{\prime(i)}), \phi(\bm{x}_t^{\prime(j)}))}{\sum_{l=1}^{|\bm{\mathcal{X}}'_t|} \sum_{m=1}^{|\bm{\mathcal{X}}'_t|} \mathbbm{1}_{c_1c_2}(y_t^{\prime(l)}, y_t^{\prime(m)})} \\
    &- 2 \sum_{i=1}^{|\bm{\mathcal{X}}_s|} \sum_{j=1}^{|\bm{\mathcal{X}}'_t|} \frac{\mathbbm{1}_{c_1c_2}(y_s^{(i)}, y_t^{\prime(j)})k(\phi(\bm{x}_s^{(i)}), \phi(\bm{x}_t^{\prime(j)}))}{\sum_{l=1}^{|\bm{\mathcal{X}}_s|} \sum_{m=1}^{|\bm{\mathcal{X}}'_t|} \mathbbm{1}_{c_1c_2}(y_s^{(l)}, y_t'^{(m)})},
  \end{aligned}
\end{equation}
with $\mathbbm{1}_{c_1c_2}(y_1, y_2)=\left\{
\begin{aligned}
1, & \, \mathrm{if} \, y_1=c_1, y_2=c_2, \\
0, & \, \mathrm{else}.
\end{aligned}
\right.$ and $c_1$ and $c_2$ representing two classes. If $c_1$ and $c_2$ are the same class, then $\mathcal{D}_{\mathrm{c_1c_2}}^{\mathrm{MMD}}$ estimates the intra-class discrepancy between the source and target domain. If $c_1$ and $c_2$ represent two different classes, then $\mathcal{D}_{\mathrm{c_1c_2}}^{\mathrm{MMD}}$ evaluates the inter-class discrepancy between both domains.

\begin{figure}[h]
\centering
\includegraphics[width=0.9\linewidth]{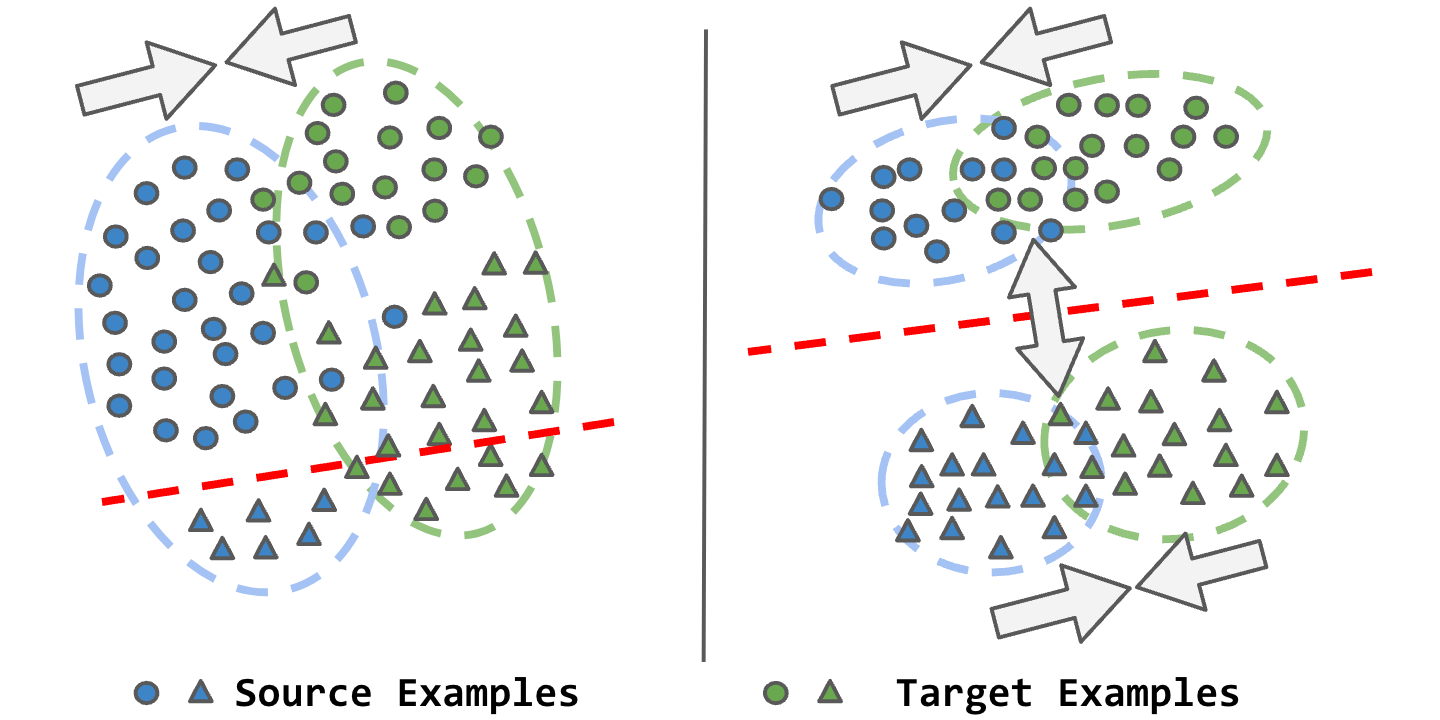}
\caption{Comparison between adapting marginal distributions (left) and conditional distributions (right). Combined with label correction and class-aware sampling, we improve the estimation of the joint distribution in the target domain.} 
\label{fig:marginal-conditional}
\vspace{-7pt}
\end{figure}

Using the class-aware MMD distance, we define the contrastive loss $\mathcal{L}_{\mathrm{contrasive}}$ as a part of the optimization objective:
\begin{equation}
  \mathcal{L}_{\mathrm{contrasive}} = \mathcal{D}_{00}^{\mathrm{MMD}} + \mathcal{D}_{11}^{\mathrm{MMD}} - \frac{1}{2} (\mathcal{D}_{01}^{\mathrm{MMD}} + \mathcal{D}_{10}^{\mathrm{MMD}}).
\end{equation}
As we consider binary classification in misinformation detection, $\mathcal{D}_{00}^{\mathrm{MMD}}$ and $\mathcal{D}_{11}^{\mathrm{MMD}}$ refers to the intra-class discrepancy of misinformation examples and credible examples in the representation space. $\mathcal{D}_{01}^{\mathrm{MMD}} + \mathcal{D}_{10}^{\mathrm{MMD}}$ represents inter-class discrepancy, the distance between different classes is enlarged to avoid misclassification (by taking the negative value). Based on the combination of intra-class and inter-class discrepancy, the misinformation detection system $\bm{f}$ learns a feature representation that separates the input from different classes and confuses input from the same class. In other words, the loss pulls together the same-class samples and pushes apart samples from different classes. As such, we minimize the distance between $p(\bm{x}|y)$ and $q(\bm{x}|y)$ based on the pseudo labels generated in the previous stage, we illustrate the adaptation process in \Cref{fig:marginal-conditional}. Existing methods (left) that adapts marginal distributions can cause performance drops when there exist label shifts between the source and target domains (see green triangles). As opposed to such methods, \ours (right) corrects the label shift and samples from the target distribution to adapt the conditional distributions, which yields an improved decision boundary for the target domain.

By combining both the NLL loss for classification and the above contrastive loss, we formulate the overall optimization objective in our contrastive adaptation stage as follows:
\begin{equation}
  \mathcal{L} = \mathcal{L}_{\mathrm{nll}} + \lambda \mathcal{L}_{\mathrm{contrasive}},
  \label{eq:overall-loss}
\end{equation}
where $\mathcal{L}_{\mathrm{nll}}$ incorporates knowledge from both the source and target domains (via pseudo labels), while $\mathcal{L}_{\mathrm{contrasive}}$ adapts the conditional distribution to the target domain. $\lambda$ can be chosen empirically.

\begin{algorithm}[t]
    \caption{Optimization loop of the proposed \ours}
    \label{alg:optimization}
    \SetAlgoLined
    \textbf{Input}
    Source examples $\bm{\mathcal{X}}_s$, unlabeled target examples $\bm{\mathcal{X}}_t$, pretrained model $\bm{f}$, correction parameters $\bm{w}$ and $\bm{b}$\;
    \textbf{Input}
    Number of iteration $N$, batch size $B$, confidence threshold $\tau$, scaling factor $\lambda$\;
    
    \tcp{1. Pseudo Labeling and Label Correction}
    Compute the target examples with pseudo labels $\bm{\mathcal{X}}'_t$\;
    Optimize the parameters $\bm{w}$ and $\bm{b}$ with \Cref{eq:label-correction-optimization}\;
    Correct the pseudo labels with \Cref{eq:label-correction}\;
    Filter $\bm{\mathcal{X}}'_t$ with minimum confidence threshold $\tau$\;
    
    \tcp{2. Contrastive Adaptation}
    \For{$i \in \{ 1, 2, ..., N \}$}{
        Sample a target batch $\{ (\bm{x}_t^{\prime(1)}, y_t^{\prime(1)}), ..., (\bm{x}_t^{\prime(B)}, y_t^{\prime(B)}) \}$\;
        Perform class-aware sampling to build a source batch $\{ (\bm{x}_s^{(1)}, y_s^{(1)}), ..., (\bm{x}_s^{(B)}, y_s^{(B)}) \}$\;
        Compute the loss with \Cref{eq:overall-loss}\;
        Perform backpropagation and update $\bm{f}$\;
    }
\end{algorithm}

\subsection{Overall Framework}
The overall framework of \ours can be found in \Cref{fig:method}. In the pseudo labeling and label correction stage, pseudo labels are first generated for all target examples using the pretrained misinformation detection system $\bm{f}$. Next, we optimize the learnable correction parameters and correct the generated pseudo labels, we also filter the target data with a minimum confidence threshold $\tau$ (chosen empirically). In the contrastive adaptation stage, we perform class-aware sampling to resemble the target data distribution. Then, we compute a contrastive adaptation loss to reduce the intra-class discrepancy and enlarge the inter-class discrepancy. For each epoch, the optimization process is described in \Cref{alg:optimization}.

Different from previous domain adaptation work in text classification~\cite{du-etal-2020-adversarial, li2021multi, zou-etal-2021-unsupervised}, we discard domain-adversarial training for efficient training and approximate the joint distribution of the target domain. In particular, we propose a label correction component to correct the estimation of $q(y)$ during pseudo labeling. This corresponds the correction of pseudo labels in COVID misinformation, where we remove the potential bias in the pretrained model to reduce noisy labels in adaptation. Next, the model is adapted by minimizing the distance between the conditional distributions via MMD distance, in which the model learns domain-confusing yet class-separating features. Put simply, we transfer the knowledge from the source misinformation data to the COVID domain via fine-grained alignment in the feature space. By estimating $q(y)$ and adapting $p(\bm{x}|y) \approx q(\bm{x}|y)$, we improve the estimation of $q(\bm{x}, y)$ with $q(\bm{x}, y) = p(\bm{x}|y)q(y)$ and thereby improving the adaptation performance for misinformation detection in COVID-19.
\section{Experiments}
\label{sec:experiment}

\begin{table}[t]
\small
\centering
\begin{tabular}{@{}lcccc@{}}
\toprule
\textbf{Datasets}    & \textbf{Negative} & \textbf{Positive} & \textbf{Avg. Len} & \textbf{Type} \\ \midrule
\multicolumn{5}{c}{\textbf{Source Datasets}}                                                     \\ \midrule
\textbf{FEVER}       & 29.59\%           & 70.41\%           & 9.4               & Claim         \\
\textbf{GettingReal} & 8.79\%            & 91.21\%           & 738.9             & News          \\
\textbf{GossipCop}   & 24.19\%           & 75.81\%           & 712.9             & News          \\
\textbf{LIAR}        & 44.23\%           & 55.77\%           & 20.2              & Claim         \\
\textbf{PHEME}       & 33.99\%           & 66.01\%           & 21.5              & Social Media  \\ \midrule
\multicolumn{5}{c}{\textbf{Target Datasets}}                                                     \\ \midrule
\textbf{CoAID}       & 9.72\%            & 90.28\%           & 54.0              & News / Claim  \\
\textbf{Constraint}  & 47.66\%           & 52.34\%           & 32.7              & Social Media  \\
\textbf{ANTiVax}     & 38.15             & 61.85\%           & 26.2              & Social Media  \\ \bottomrule
\end{tabular}
\caption{Dataset details.}
\label{tab:datasets}
\vspace{-7pt}
\end{table}

\subsection{Experimental Design}
\textbf{Datasets}: In our experiments, we adopt five source datasets released before the COVID outbreak and three COVID misinformation datasets to validate the proposed \ours. For source datasets, we use FEVER~\cite{thorne2018fever}, GettingReal~\cite{meganrisdal2016}, GossipCop~\cite{shu2020fakenewsnet}, LIAR~\cite{wang2017liar} and PHEME~\cite{buntain2017automatically}. For target datasets, we adopt CoAID~\cite{cui2020coaid}, Constraint~\cite{patwa2021fighting} and ANTiVax~\cite{hayawi2022anti}, which are released in 2020, 2021 and 2022 respectively. Dataset details are provided in \Cref{tab:datasets}, where \textbf{Negative} and \textbf{Positive} are the proportion of misinformation and valid information in the dataset. \textbf{Avg. Len} represents the average token length of text and \textbf{Type} denotes the source type of the text (e.g., news, claim or social media).

\textbf{Model}: Following~\cite{li2021multi, zou-etal-2021-unsupervised, liu2019roberta}, we select the commonly used RoBERTa as the misinformation detection model to perform the proposed \ours. RoBERTa is a transformer-based language model with modified pretraining method that improves the performance on various NLP downstream tasks.

\textbf{Baselines}: The pretrained misinformation detection model is first directly evaluated on the target test data as na{\"i}ve baseline. We additionally select two state-of-the-art baselines: DAAT and EADA~\cite{du-etal-2020-adversarial, zou-etal-2021-unsupervised}. DAAT leverages post training on BERT to improve the domain-adversarial adaptation. EADA performs energy-based adversarial domain adaptation with an additional autoencoder.

\textbf{Evaluation}: Similar to~\cite{kou2022hc, li2021multi, zou-etal-2021-unsupervised}, we split the data into training, validation, and test sets with the ratio of 7:1:2. We adopt accuracy and F1 score to perform evaluation. We additionally introduce balanced accuracy (BA) to equivalently evaluate the adaptation performance in both classes, balanced accuracy is defined as the average value of sensitivity and specificity:
\begin{equation}
  \mathrm{BA} = \frac{1}{2} (\mathrm{TPR} + \mathrm{TNR}) = \frac{1}{2} (\frac{\mathrm{TP}}{\mathrm{TP} + \mathrm{FN}} + \frac{\mathrm{TN}}{\mathrm{TN} + \mathrm{FP}}),
\end{equation}
where TPR represents sensitivity and TNR represents specificity. TP, TN represent true positive and true negative, FP and FN represent false positive and false negative. In other words, balanced accuracy equally considers the accuracy of both classes, and thus yields a better estimation of adaptation performance under label shift.

\begin{table}[t]
\small
\centering
\begin{tabular}{@{}lccc@{}}
\toprule
\textbf{Dataset \quad} & \textbf{\quad BA $\uparrow$ \quad} & \textbf{\quad Acc. $\uparrow$ \quad} & \textbf{\quad F1 $\uparrow$ \quad} \\ \midrule
\multicolumn{4}{c}{\textbf{Source Datasets}}                                                                                            \\ \midrule
\textbf{FEVER}         & 0.7957                             & 0.7957                               & 0.8173                             \\
\textbf{GettingReal}   & 0.8459                             & 0.9587                               & 0.9776                             \\
\textbf{GossipCop}     & 0.7760                             & 0.8693                               & 0.9173                             \\
\textbf{LIAR}          & 0.6070                             & 0.6322                               & 0.7116                             \\
\textbf{PHEME}         & 0.8630                             & 0.8674                               & 0.8975                             \\ \midrule
\multicolumn{4}{c}{\textbf{Target Datasets}}                                                                                            \\ \midrule
\textbf{CoAID}         & 0.8892                             & 0.9720                               & 0.9846                             \\
\textbf{Constraint}    & 0.9350                             & 0.9327                               & 0.9323                             \\
\textbf{ANTiVax}       & 0.9303                             & 0.9191                               & 0.9291                             \\ \bottomrule
\end{tabular}
\caption{Supervised training results.}
\label{tab:supervised}
\vspace{-7pt}
\end{table}

\begin{table*}[t]
\small
\centering
\begin{tabular}{@{}lcccccccccc@{}}
\toprule
\multirow{2}{*}{\textbf{Source Dataset}} & \textbf{Target Dataset} & \multicolumn{3}{c}{\textbf{CoAID (2020)}}           & \multicolumn{3}{c}{\textbf{Constraint (2021)}}      & \multicolumn{3}{c}{\textbf{ANTiVax (2022)}}         \\ \cmidrule(l){2-11} 
                                         & Metric                  & BA $\uparrow$   & Acc. $\uparrow$ & F1 $\uparrow$   & BA $\uparrow$   & Acc. $\uparrow$ & F1 $\uparrow$   & BA $\uparrow$   & Acc. $\uparrow$ & F1 $\uparrow$   \\ \midrule
\multirow{4}{*}{FEVER (2018)}            & None                    & \textbf{0.5528} & \ul{0.8980}     & \ul{0.9456}     & \ul{0.5633}    & \ul{0.5818}      & \textbf{0.7057} & \ul{0.5871}     & 0.6219          & 0.7066          \\
                                         & DAAT                    & 0.5113          & 0.8920          & 0.9427          & 0.5485          & 0.5640          & 0.6791          & 0.5641          & \ul{0.6255}     & 0.7345          \\
                                         & EADA                    & 0.5185          & \textbf{0.9050} & \textbf{0.9499} & 0.5602          & 0.5780          & \ul{0.7001}     & 0.5223          & 0.6159          & \textbf{0.7531} \\
                                         & CANMD                   & \ul{0.5236}     & 0.8970          & 0.9454          & \textbf{0.6837} & \textbf{0.6827} & 0.6864          & \textbf{0.6501} & \textbf{0.6793} & \ul{0.7485}     \\ \midrule
\multirow{4}{*}{GettingReal (2016)}      & None                    & 0.5000          & 0.9060          & 0.9507          & 0.5132          & 0.5360          & 0.6929          & 0.5022          & 0.6066          & \ul{0.7546}     \\
                                         & DAAT                    & \ul{0.5314}     & \ul{0.9110}     & \ul{0.9531}    & \ul{0.5532}      & \ul{0.5734}     & 0.7075          & \ul{0.5057}     & \ul{0.6093}     & \textbf{0.7557} \\
                                         & EADA                    & 0.5000          & 0.9060          & 0.9507          & 0.5466          & 0.5678          & \ul{0.7077}     & 0.5025          & 0.6070          & 0.7548          \\
                                         & CANMD                   & \textbf{0.5622} & \textbf{0.9150} & \textbf{0.9551} & \textbf{0.7440} & \textbf{0.7416} & \textbf{0.7370} & \textbf{0.5629} & \textbf{0.6289} & 0.7409          \\ \midrule
\multirow{4}{*}{GossipCop (2018)}        & None                    & 0.5121          & 0.9020          & 0.9483          & 0.5559          & 0.5762          & 0.7096          & 0.5283          & 0.6262          & 0.7630          \\
                                         & DAAT                    & \ul{0.5095}     & \ul{0.9060}     & \ul{0.9506}     & \ul{0.7178}     & \ul{0.7276}     & \ul{0.7806}     & \ul{0.6692}     & 0.7161          & \ul{0.7918}     \\
                                         & EADA                    & 0.4994          & 0.9050          & 0.9501          & 0.5210          & 0.5430          & 0.6944          & 0.5509          & \ul{0.6434}     & 0.7709          \\
                                         & CANMD                   & \textbf{0.6487} & \textbf{0.9250} & \textbf{0.9598} & \textbf{0.7914} & \textbf{0.7921} & \textbf{0.8020} & \textbf{0.7410} & \textbf{0.7771} & \textbf{0.8321} \\ \midrule
\multirow{4}{*}{LIAR  (2017)}            & None                    & 0.6900          & 0.7320          & 0.8337          & 0.7255          & 0.7318          & 0.7706          & 0.5147          & 0.4846          & 0.4656          \\
                                         & DAAT                    & 0.5000          & 0.9060          & 0.9507          & 0.7607          & 0.7626          & 0.7795          & \ul{0.6228}     & 0.5778          & 0.5393          \\
                                         & EADA                    & \ul{0.7019}     & \ul{0.9350}     & \ul{0.9650}     & \ul{0.7776}     & \ul{0.7794}     & \ul{0.7950}     & 0.6184          & \ul{0.6642}     & \ul{0.7490}     \\
                                         & CANMD                   & \textbf{0.7164} & \textbf{0.9440} & \textbf{0.9699} & \textbf{0.8147} & \textbf{0.8140} & \textbf{0.8183} & \textbf{0.7554} & \textbf{0.7844} & \textbf{0.8338} \\ \midrule
\multirow{4}{*}{PHEME  (2017)}           & None                    & 0.5015          & 0.9000          & 0.9473          & 0.4787          & 0.4972          & \ul{0.6456}     & 0.5083          & 0.6103          & 0.7553          \\
                                         & DAAT                    & \ul{0.5149}     & \textbf{0.9070} & \textbf{0.9511} & \ul{0.5227}     & \ul{0.5411}     & \textbf{0.6763} & \ul{0.5895}     & \ul{0.6498}     & 0.7518          \\
                                         & EADA                    & 0.5095          & \ul{0.9060}     & \ul{0.9506}     & 0.4944          & 0.4969          & 0.6391          & 0.5411          & 0.6328          & \ul{0.7632}     \\
                                         & CANMD                   & \textbf{0.5152} & 0.8990          & 0.9466          & \textbf{0.5594} & \textbf{0.5654} & 0.6235          & \textbf{0.6166} & \textbf{0.6823} & \textbf{0.7797} \\ \midrule
\multirow{4}{*}{Average}                 & None                    & \ul{0.5513}     & 0.8676          & 0.9251          & 0.5673          & 0.5846          & 0.7049          & 0.5281          & 0.5899          & 0.6890          \\
                                         & DAAT                    & 0.5134          & 0.9044          & 0.9496          & \ul{0.6206}     & \ul{0.6337}     & \ul{0.7246}     & \ul{0.5903}     & \ul{0.6357}     & 0.7146          \\
                                         & EADA                    & 0.5458          & \ul{0.9114}     & \ul{0.9533}     & 0.5800          & 0.5930          & 0.7073          & 0.5470          & 0.6327          & \ul{0.7582}     \\
                                         & CANMD                   & \textbf{0.5932} & \textbf{0.9160} & \textbf{0.9554} & \textbf{0.7186} & \textbf{0.7192} & \textbf{0.7334} & \textbf{0.6652} & \textbf{0.7104} & \textbf{0.7870} \\ \bottomrule
\end{tabular}
\caption{Main results of domain adaptation.}
\label{tab:results}
\vspace{-7pt}
\end{table*}

\textbf{Implementation}: The input text is preprocessed as follows: (1)~special symbols like emojis are back translated into English words; (2)~hashtags, mentions and URLs are tokenized for posts on social media; (3)~special characters are removed from the input text. For training, we first pretrain the misinformation detection model on the source dataset using Adam optimizer without linear warm-up. We adopt the learning rate of 1e-5, batch size of 24 and the maximum epoch of 5. We validate the model after each epoch with the validation set and report metric scores on the test set. In the adaptation step of \ours, we adopt the same training and evaluation methods. For batching, we sample 24 target examples and perform class-aware sampling to sample another 24 source examples. We select confidence threshold $\tau$ from $0.6$ to $0.8$ and scaling factor $\lambda$ from $0.001$ to $0.05$. For baseline methods, we adopt the identical training conditions as \ours and follow the default hyperparameter selections in the original papers~\cite{du-etal-2020-adversarial, zou-etal-2021-unsupervised}.

\begin{figure*}[t]
\centering
	\begin{subfigure}[b]{0.19\textwidth}
      \centering
      \includegraphics[trim={0 25 0 25}, width=\textwidth]{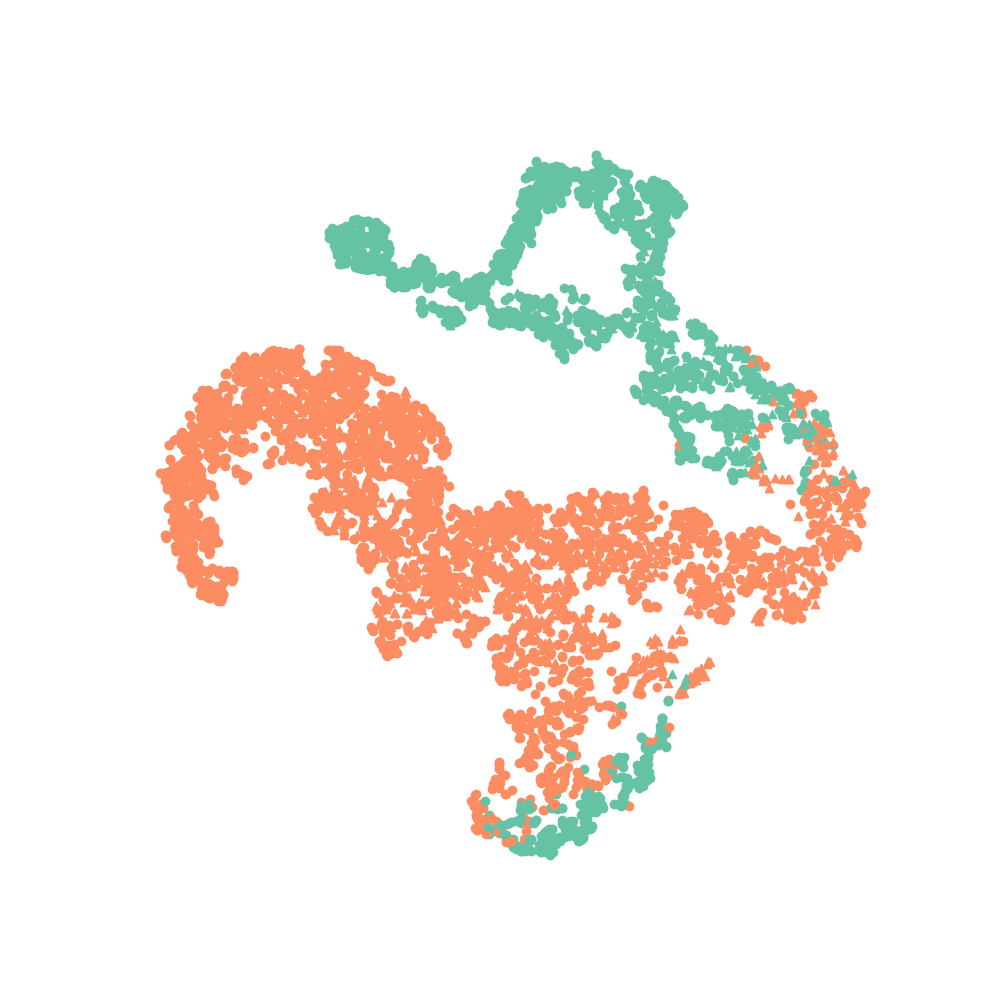}
      \caption{F $\rightarrow$ Const. (DAAT)}
	\end{subfigure}
	\begin{subfigure}[b]{0.19\textwidth}
      \centering
      \includegraphics[trim={0 25 0 25}, width=\textwidth]{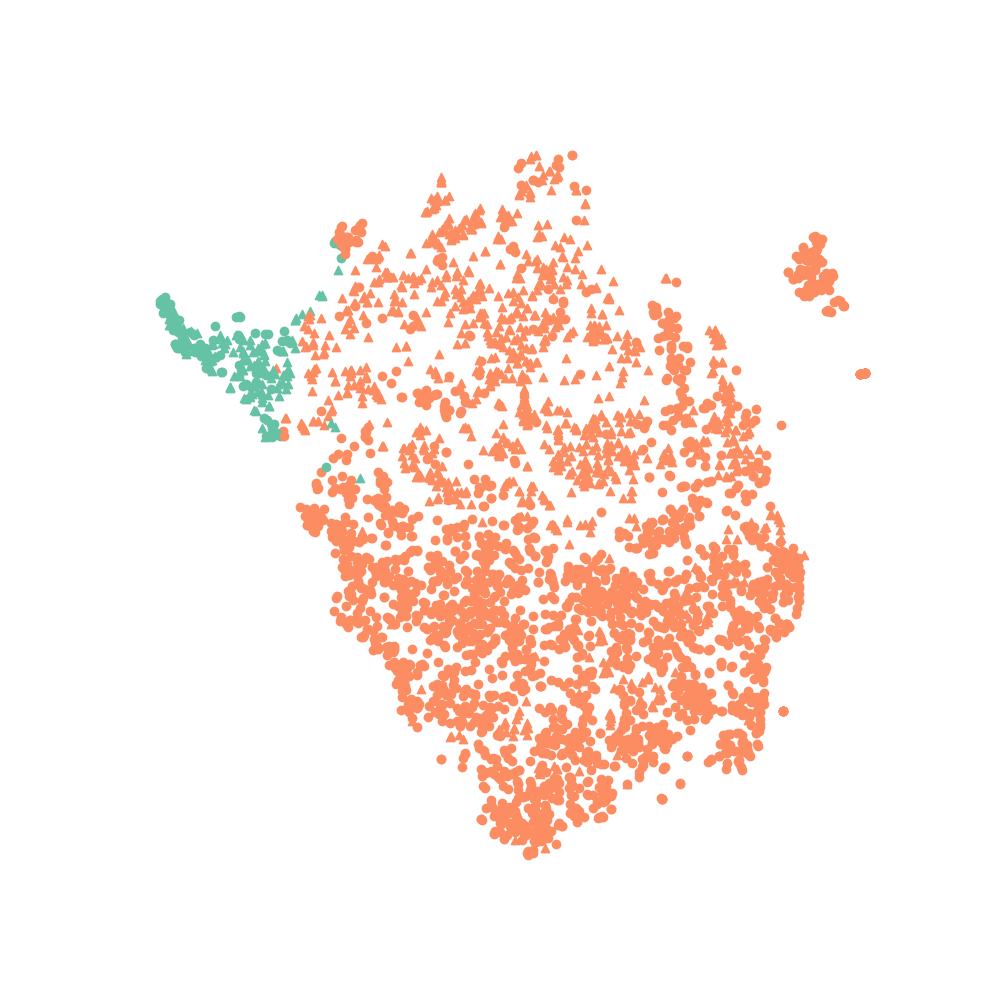}
      \caption{GR $\rightarrow$ Const. (DAAT)}
	\end{subfigure}
	\begin{subfigure}[b]{0.19\textwidth}
      \centering
      \includegraphics[trim={0 25 0 25}, width=\textwidth]{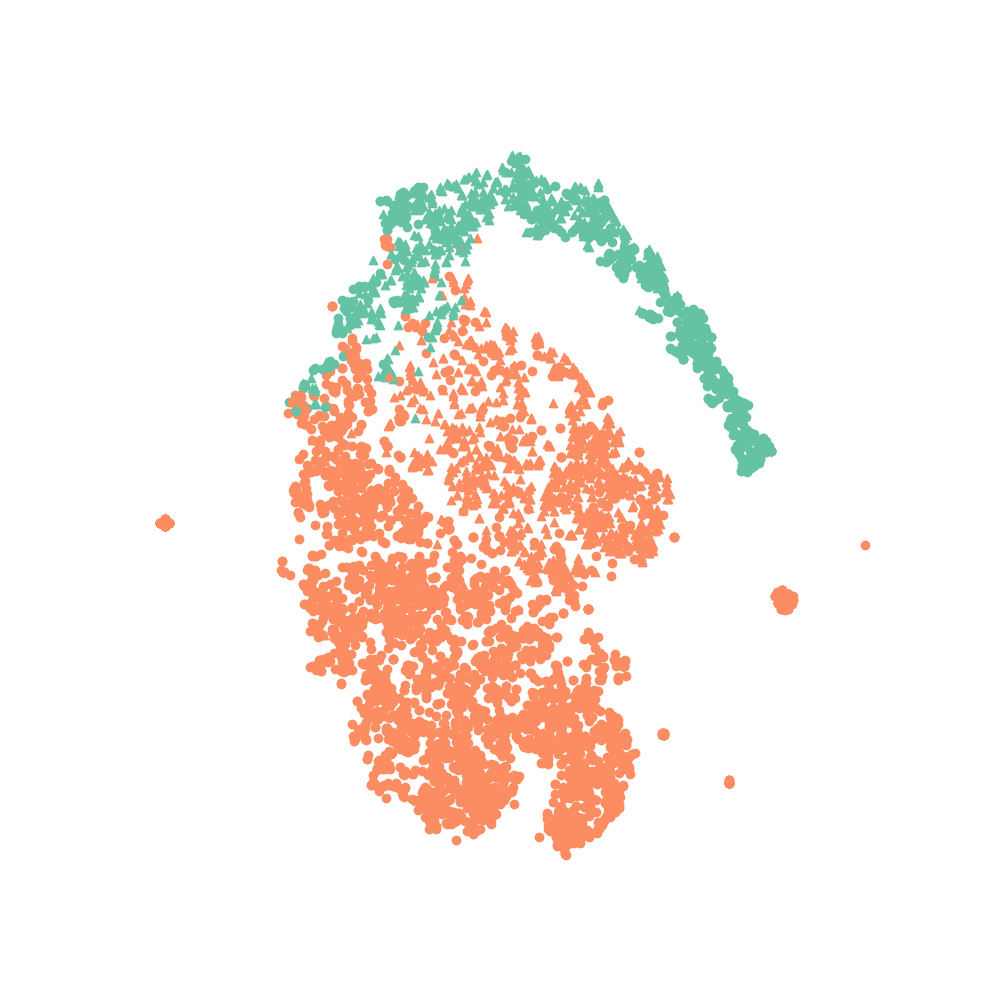}
      \caption{GC $\rightarrow$ Const. (DAAT)}
	\end{subfigure}
	\begin{subfigure}[b]{0.19\textwidth}
      \centering
      \includegraphics[trim={0 25 0 25}, width=\textwidth]{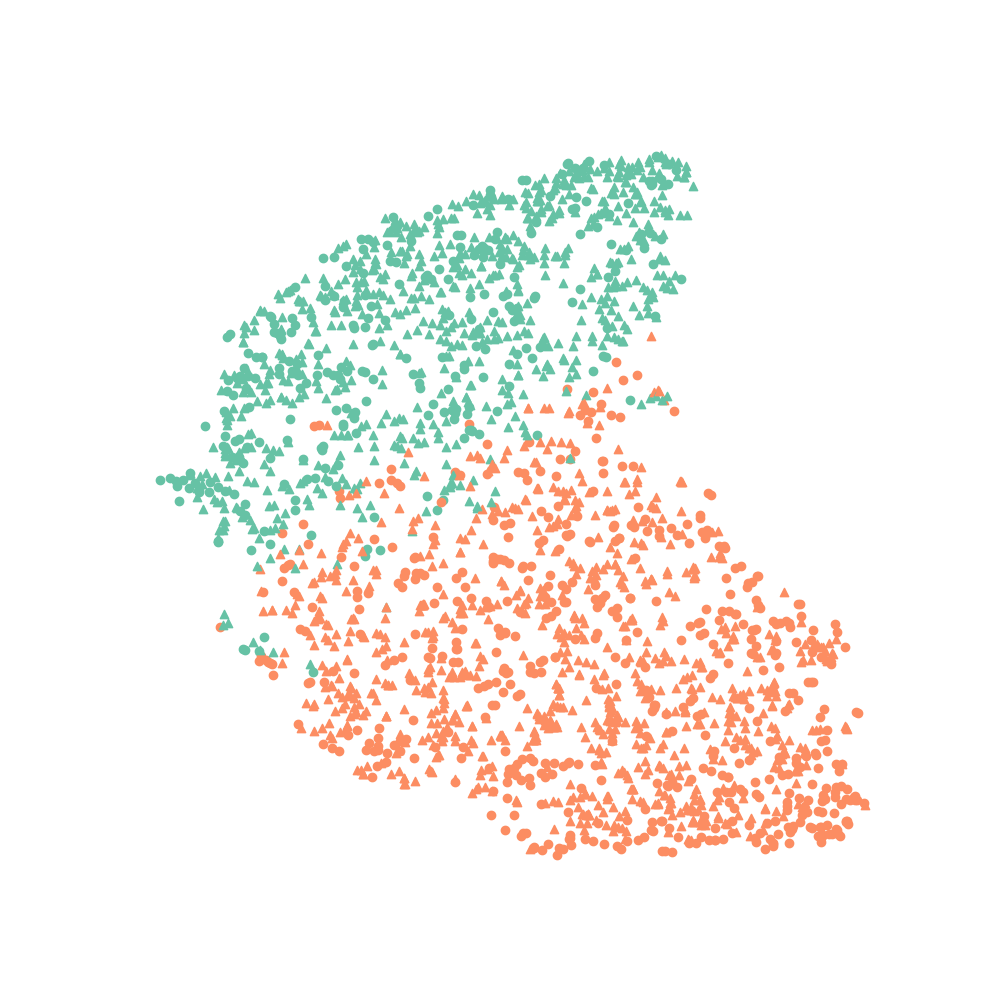}
      \caption{L $\rightarrow$ Const. (DAAT)}
	\end{subfigure}
	\begin{subfigure}[b]{0.19\textwidth}
      \centering
      \includegraphics[trim={0 25 0 25}, width=\textwidth]{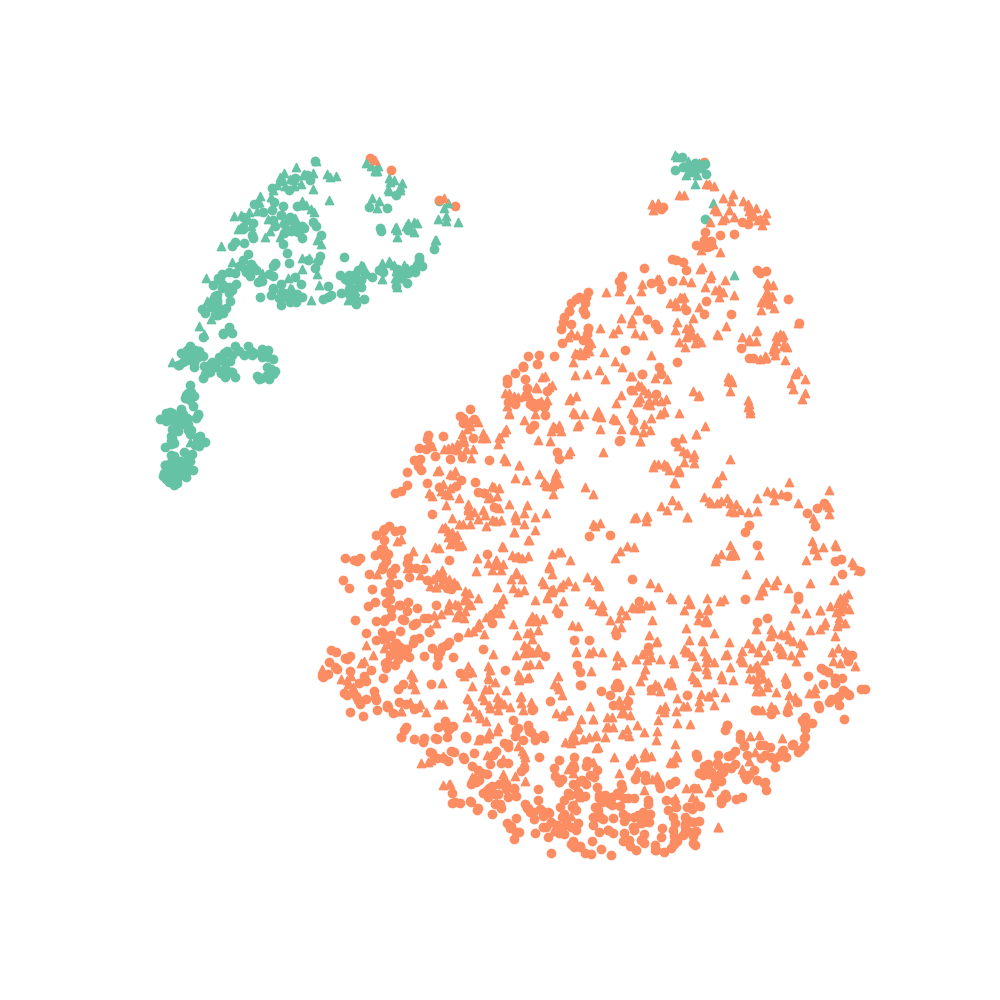}
      \caption{P $\rightarrow$ Const. (DAAT)}
	\end{subfigure}
	
	\begin{subfigure}[b]{0.19\textwidth}
      \centering
      \includegraphics[trim={0 25 0 25}, width=\textwidth]{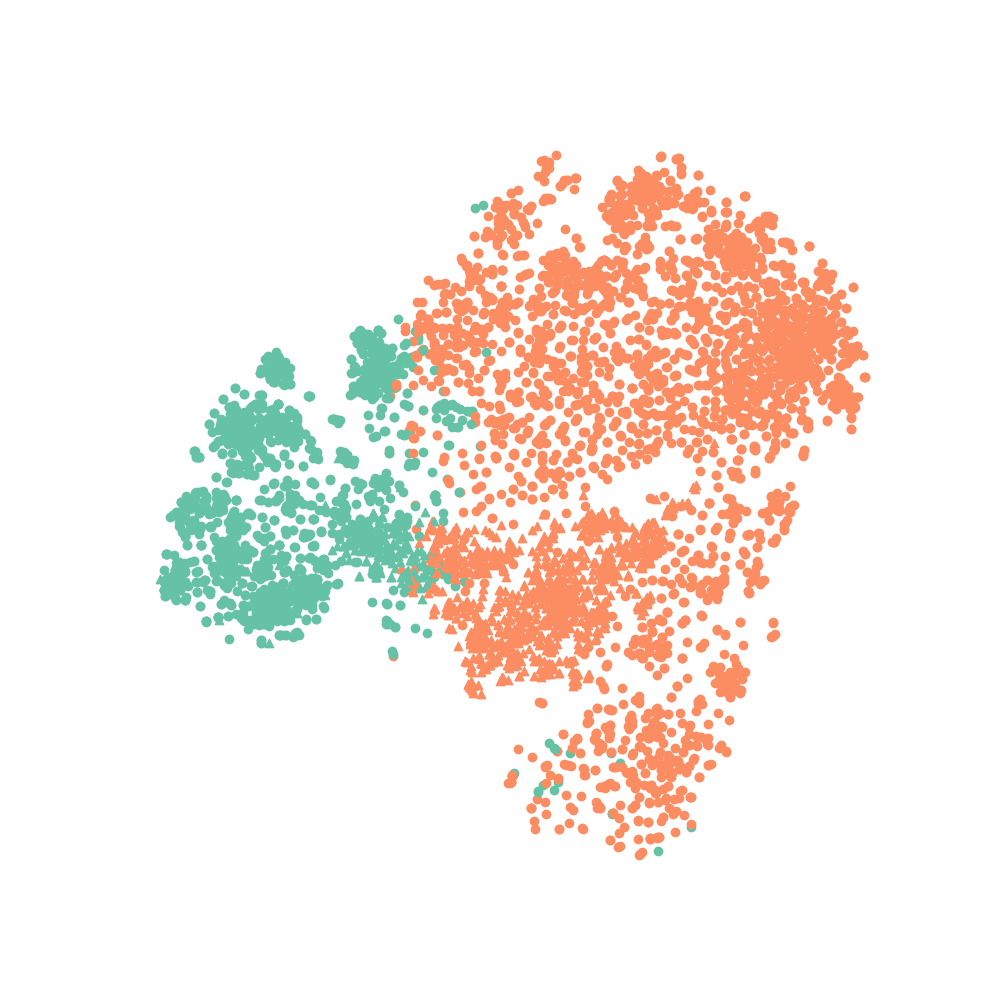}
      \caption{F $\rightarrow$ Const. (EADA)}
	\end{subfigure}
	\begin{subfigure}[b]{0.19\textwidth}
      \centering
      \includegraphics[trim={0 25 0 25}, width=\textwidth]{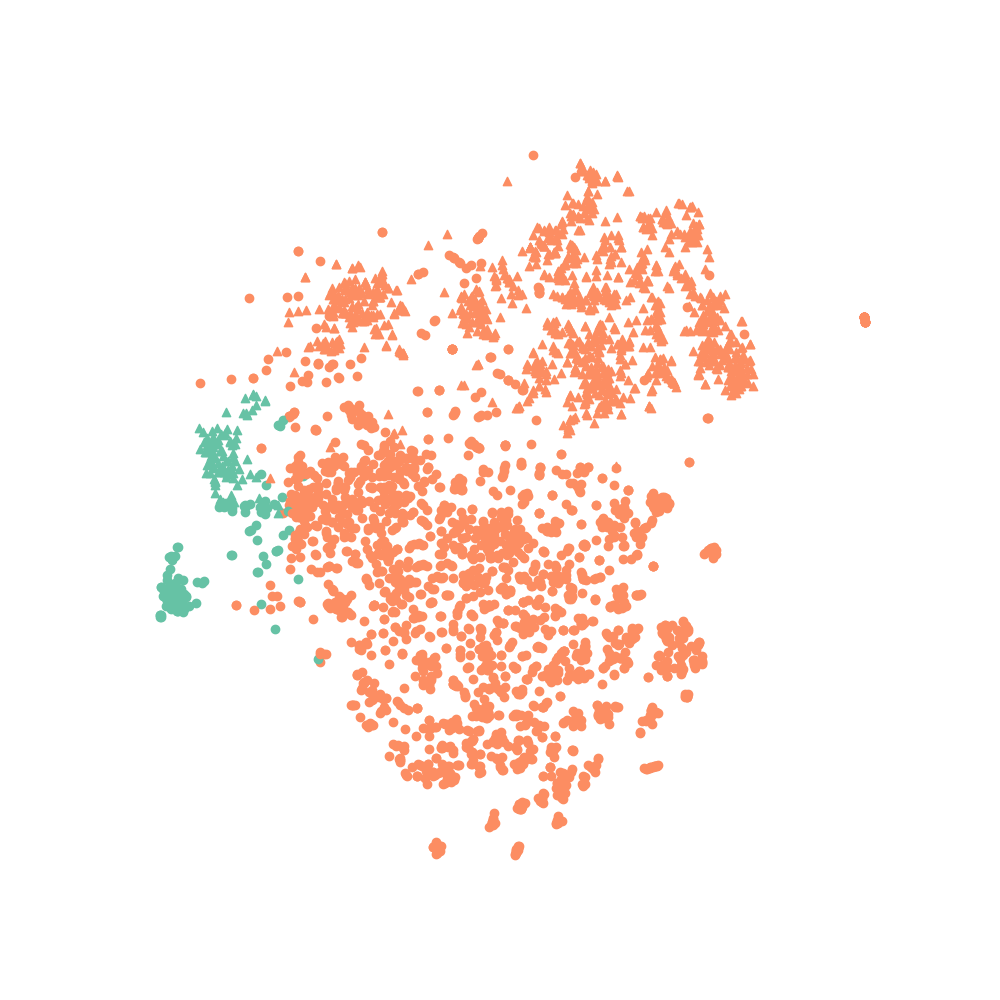}
      \caption{GR $\rightarrow$ Const. (EADA)}
	\end{subfigure}
	\begin{subfigure}[b]{0.19\textwidth}
      \centering
      \includegraphics[trim={0 25 0 25}, width=\textwidth]{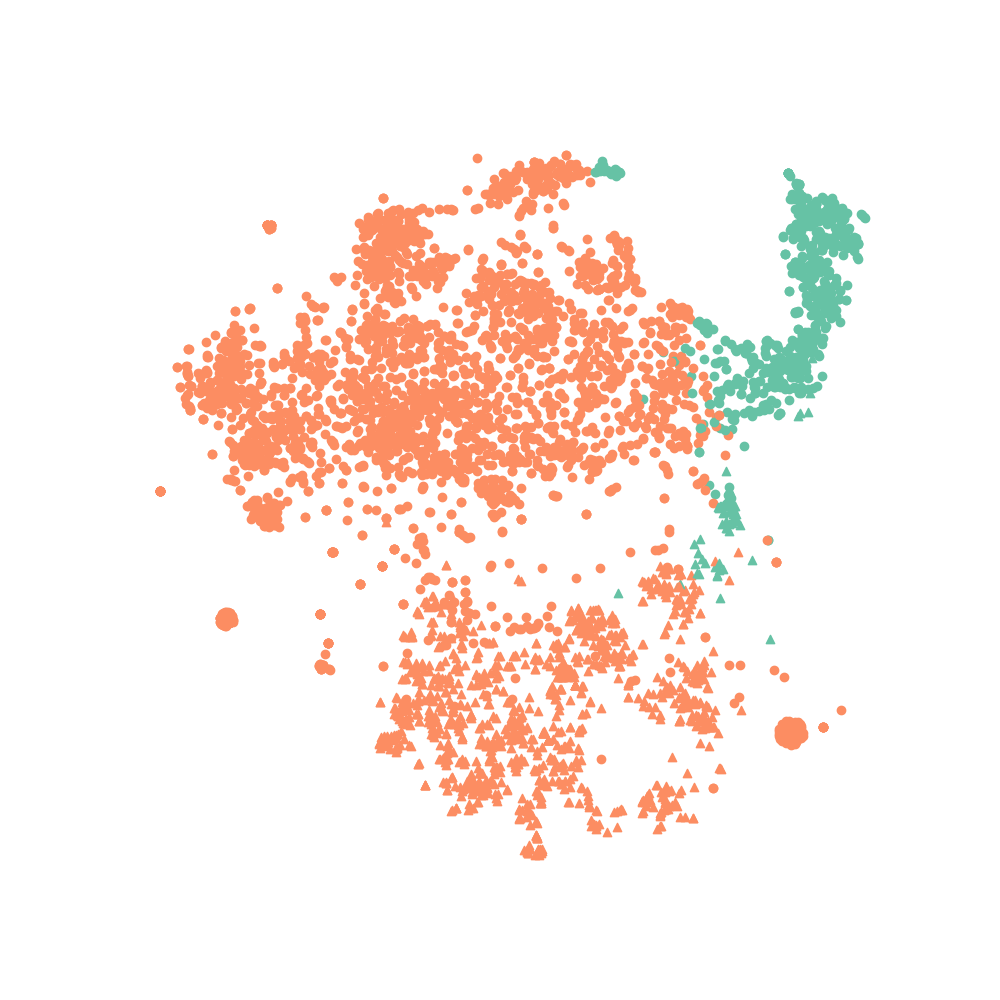}
      \caption{GC $\rightarrow$ Const. (EADA)}
	\end{subfigure}
	\begin{subfigure}[b]{0.19\textwidth}
      \centering
      \includegraphics[trim={0 25 0 25}, width=\textwidth]{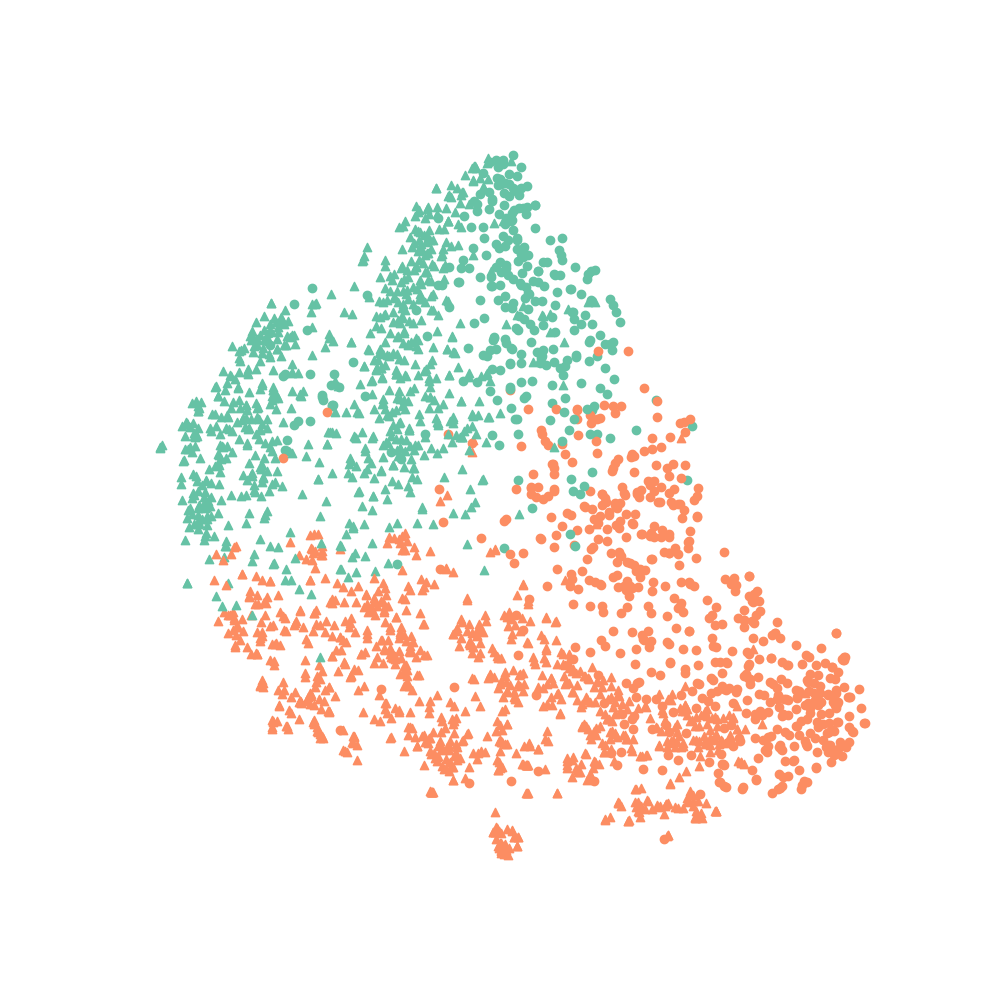}
      \caption{L $\rightarrow$ Const. (EADA)}
	\end{subfigure}
	\begin{subfigure}[b]{0.19\textwidth}
      \centering
      \includegraphics[trim={0 25 0 25}, width=\textwidth]{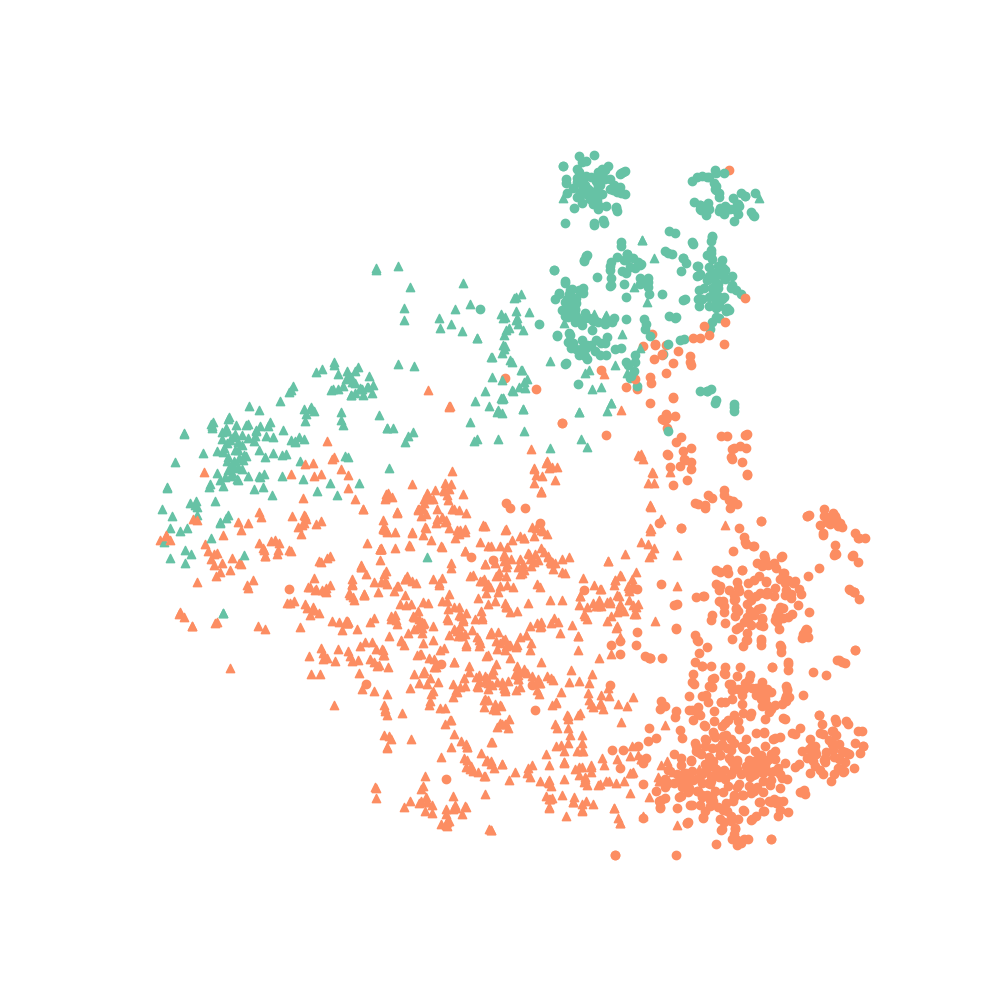}
      \caption{P $\rightarrow$ Const. (EADA)}
	\end{subfigure}
	
	\begin{subfigure}[b]{0.19\textwidth}
      \centering
      \includegraphics[trim={0 25 0 25}, width=\textwidth]{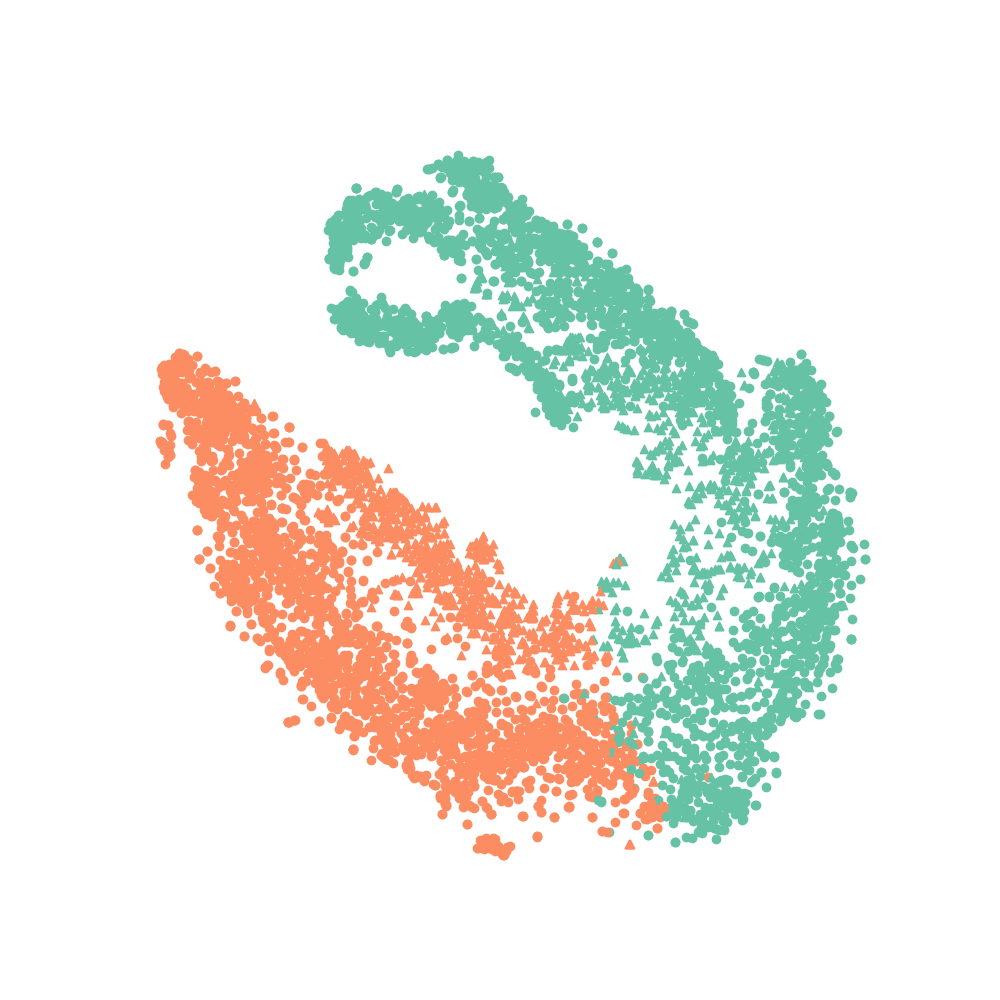}
      \caption{F $\rightarrow$ Const. (\ours)}
      \label{fig:f2c_ours}
	\end{subfigure}
	\begin{subfigure}[b]{0.19\textwidth}
      \centering
      \includegraphics[trim={0 25 0 25}, width=\textwidth]{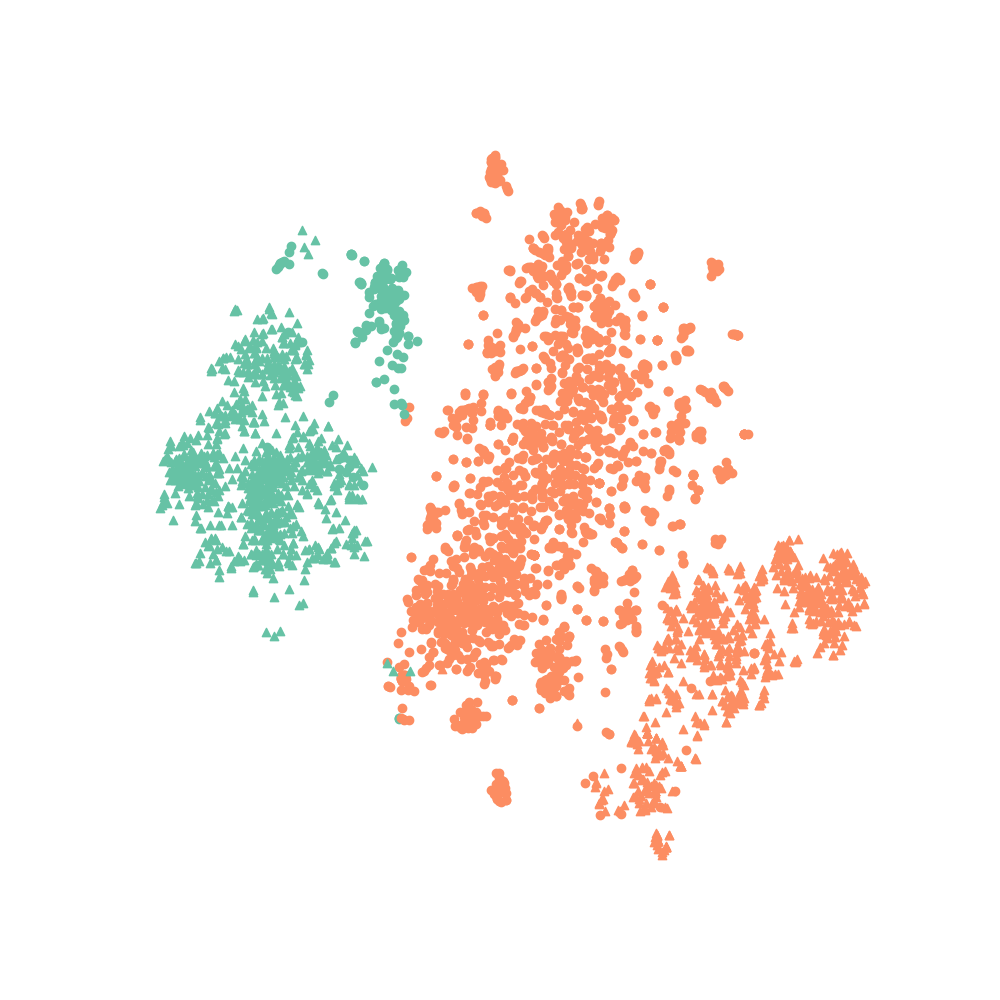}
      \caption{GR $\rightarrow$ Const. (\ours)}
      \label{fig:gr2c_ours}
	\end{subfigure}
	\begin{subfigure}[b]{0.19\textwidth}
      \centering
      \includegraphics[trim={0 25 0 25}, width=\textwidth]{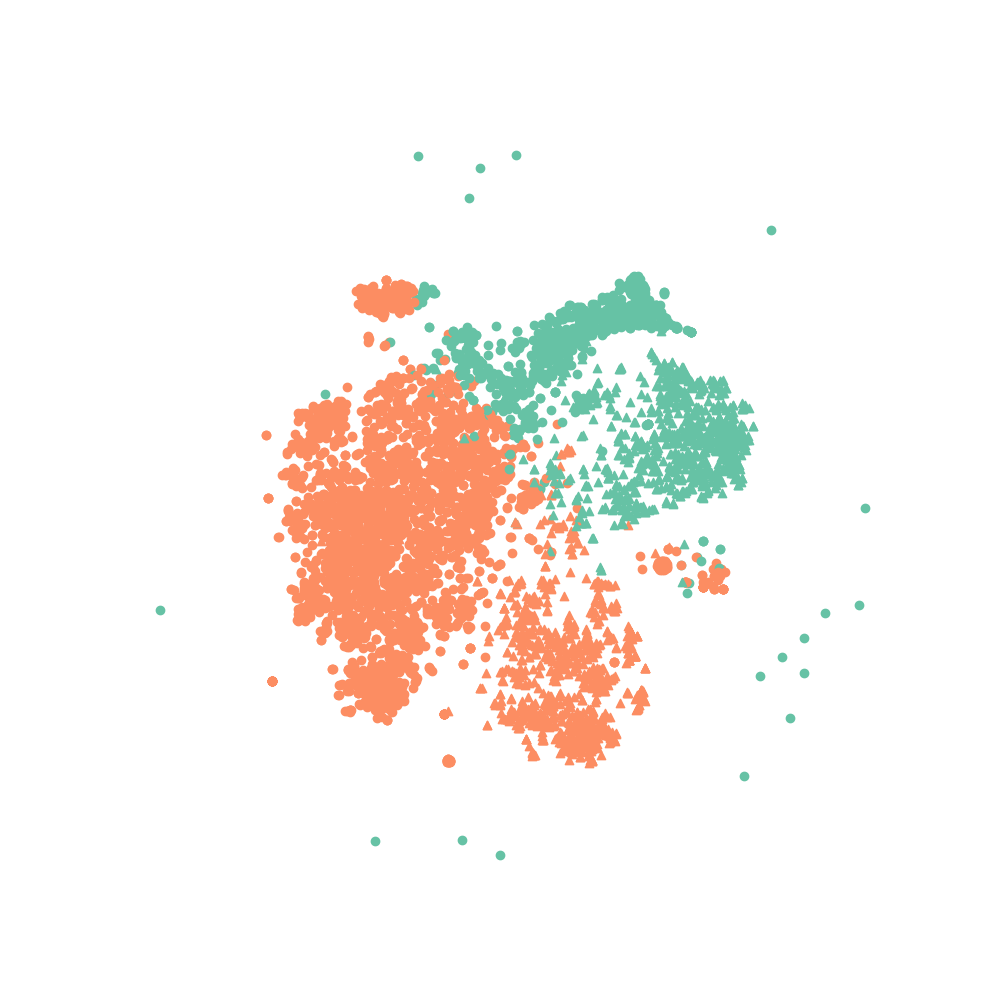}
      \caption{GC $\rightarrow$ Const. (\ours)}
      \label{fig:gc2c_ours}
	\end{subfigure}
	\begin{subfigure}[b]{0.19\textwidth}
      \centering
      \includegraphics[trim={0 25 0 25}, width=\textwidth]{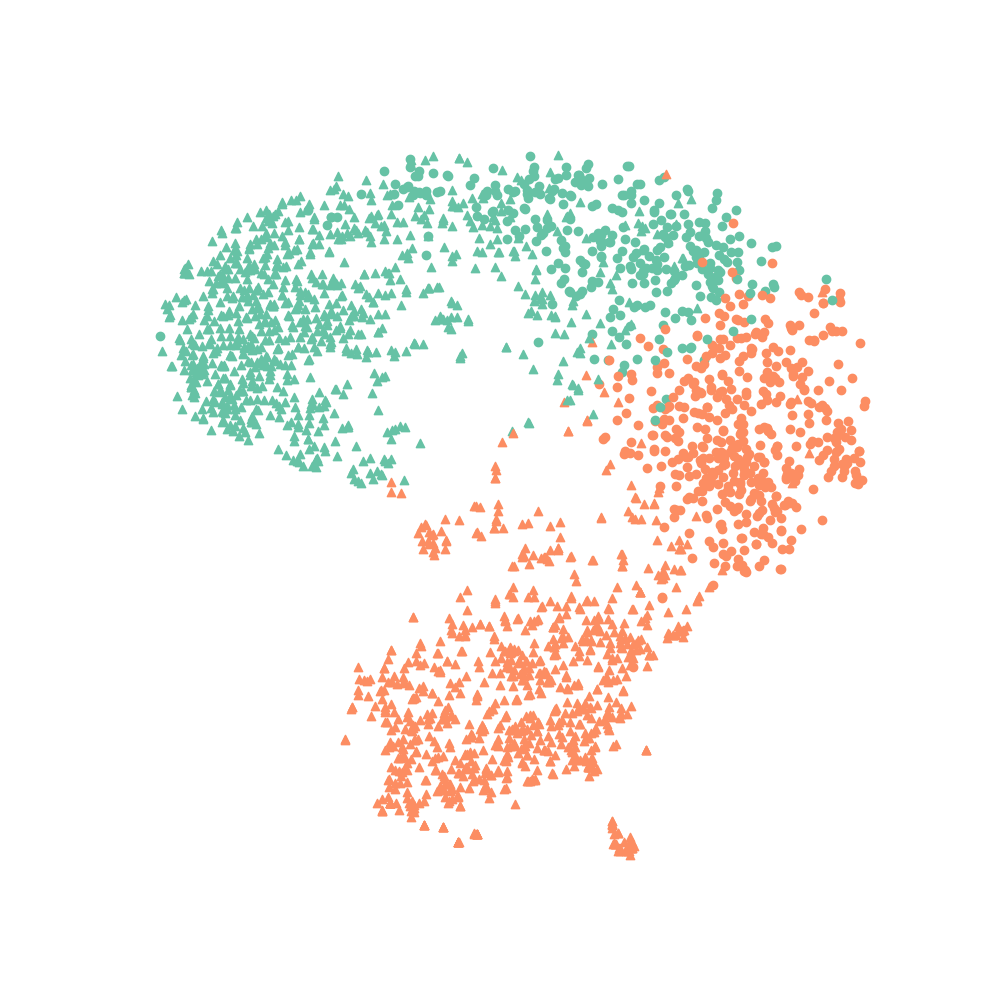}
      \caption{L $\rightarrow$ Const. (\ours)}
      \label{fig:l2c_ours}
	\end{subfigure}
	\begin{subfigure}[b]{0.19\textwidth}
      \centering
      \includegraphics[trim={0 25 0 25}, width=\textwidth]{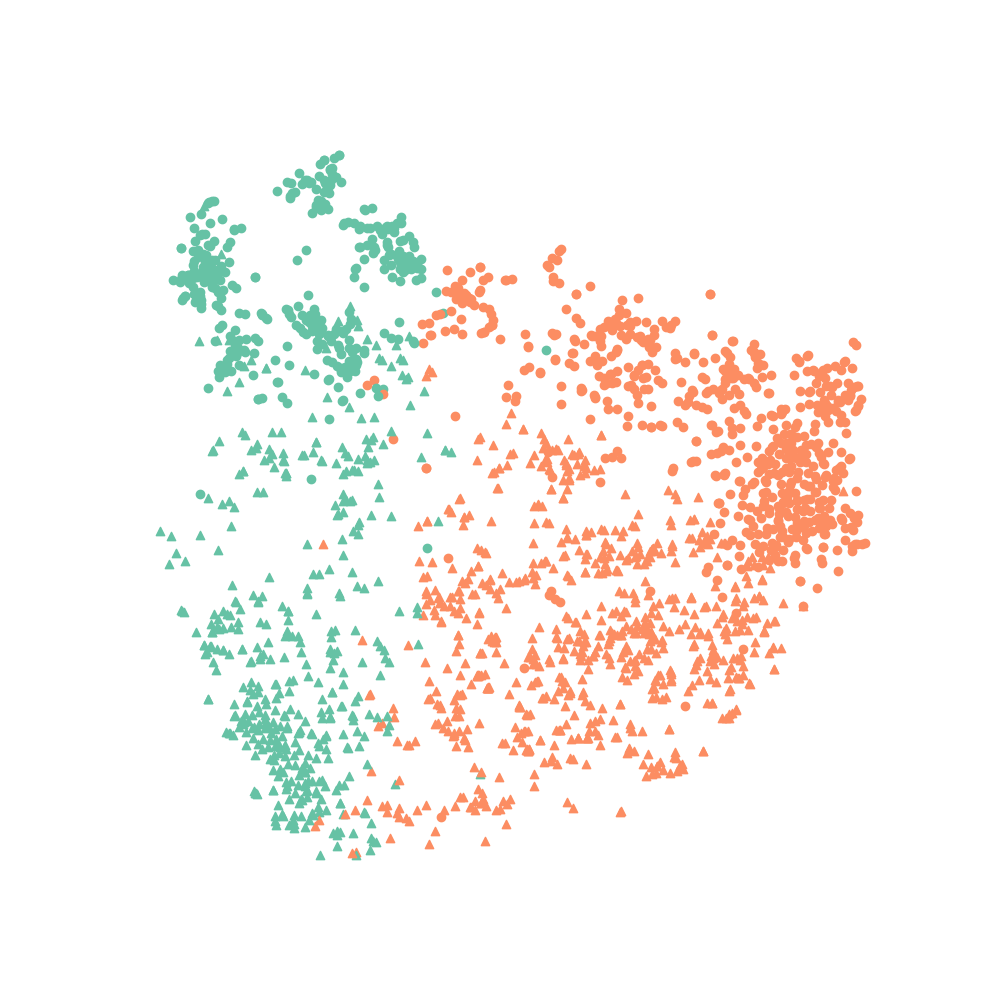}
      \caption{P $\rightarrow$ Const. (\ours)}
      \label{fig:p2c_ours}
	\end{subfigure}
\caption{Qualitative results of domain adaptation. We denote Constraint with Const., FEVER with F, GettingReal with GR, GossipCop with GC, LIAR with L and PHEME with P. True examples are marked in orange while fake examples are in green.}
\label{fig:qualitative}
\vspace{-7pt}
\end{figure*}

\subsection{Quantitative Analysis}
We first present the supervised training results of all datasets in \Cref{tab:supervised}, the evaluation results on the target datasets provide an upper bound for the adaptation performance. We observe the following: (1)~the performance varies for different data types. For example, RoBERTa does not perform well when we train on claims without additional knowledge (e.g., LIAR provides short claims). Predicting misinformation on social media posts achieves better performance with average BA of 0.9094 (i.e., PHEME, Constraint and ANTiVax), potentially due to the syntactic patterns of misinformation (e.g., second-person pronouns, swear words and adverbs~\cite{li2021multi}). (2)~For disproportionate label distributions, balanced accuracy (BA) better reflects the classification performance by weighting both classes equally. For example, the classification accuracy is 0.9587 on GettingReal, while the BA score drops drastically to 0.8459. This is because the dataset contains overwhelmingly true examples and by only predicting one class, the model still achieves over 0.9 accuracy. Overall, the supervised performance depends heavily on the input data type and label distributions, suggesting the potential benefits of domain similarity in adapting misinformation detection systems.

Now we study the adaptation performance and report the results in \Cref{tab:results}. Each row represents a combination of source datasets and domain adaptation methods, and each column includes metric scores on one target dataset. We report BA, accuracy and F1 metrics for all source-target combinations, the best results are marked in bold and the second best results are underlined. We report our observations: (1)~due to the large domain discrepancies, domain adaptation for COVID misinformation detection is a non-trivial task. For instance, the na{\"i}ve baseline (i.e., None) trained on FEVER achieves 0.5633 BA and 0.5818 accuracy in Constraint; (2)~in rare cases, we observe negative transfer. For example, consistent performance drops can be found on the BA metric in FEVER $\rightarrow$ CoAID experiments due to the discrepancy and the label shift across both domains.
(3)~Baseline adaptation methods achieve superior results than the na{\"i}ve baseline (i.e., None) in most cases. For example, the DAAT baseline achieves 4.7\% and 6.4\% relative improvements on BA and accuracy, similar improvements can be found using EADA. (4)~On average, \ours performs the best by achieving the highest performance in most scenarios and outperforming the best baseline method by 11.5\% and 7.9\% on BA and accuracy. 
(5)~\ours performs well even when the source and target datasets demonstrate significant label shifts. For instance, GettingReal consists of over 90\% true examples while Constraint is a balanced dataset in the COVID domain. \ours achieves 0.7440 BA and 0.7416 accuracy in GettingReal $\rightarrow$ Constraint, with 34.5\% and 29.3\% improvements compared to the strongest baseline (i.e., DAAT). Altogether, the results suggest that the proposed \ours is effective in improving misinformation detection performance on out-of-domain data. By comparing the performance in source-target combinations with dissimilar label distributions (e.g., GossipCop $\rightarrow$ CoAID), \ours is particularly effective in adapting models to target domains under significant label shifts due to label correction.

\subsection{Qualitative Analysis}
We also present qualitative examples of the learnt domain-invariant features generated by the baseline methods and \ours. In particular, we first train the misinformation detection models with different domain adaptation methods. The trained models are evaluated on the test set and we visualize the \texttt{[CLS]} position output from the transformer encoder. We select the correct predictions and perform T-SNE to generate the plots. The plots can be found in \Cref{fig:qualitative}. Constraint is a widely used COVID misinformation dataset, we thus perform the experiments with DAAT, EADA and \ours using Constraint as the target dataset (Const.), valid examples are in orange while misinformation examples are in green. For source datasets, we denote the FEVER with F, GettingReal with GR, GossipCop with GC, LIAR with L and PHEME with P.

From the qualitative examples we observe the following: (1)~all domain adaptation methods generate domain-invariant features in the representation space, as we do not observe clustering of examples in the plots. Nevertheless, different classes of examples are not clearly separated in most cases. (2)~The proposed \ours reduces intra-class distances while separating the true samples from the misinformation samples. For example, we observe a clear decision boundary between both classes in \Cref{fig:f2c_ours} and \Cref{fig:gr2c_ours}. (3)~The label correction component in \ours effectively corrects the feature distribution in the representation space. For instance, the target dataset Constraint has a balanced label distribution, while FEVER, GettingReal and GossipCop comprise of over 70\% credible examples (orange). Yet we observe comparable area consisting of credible and fake examples in \Cref{fig:f2c_ours}, \Cref{fig:gr2c_ours} and \Cref{fig:gc2c_ours}, whereas the baseline methods demonstrate overwhelming orange area (as of the source data distribution). In sum, \ours exploits the source domain knowledge by enlarging inter-class margins and learning compact intra-class features, \ours also corrects the label shifts to remove potential bias from the source data.

\subsection{Performance Variations}
We now discuss the potential reasons for the performance variations in the adaptation results. For this purpose, we refer to the adaptation results in \Cref{tab:results} and qualitative examples in \Cref{fig:qualitative}. In particular, we discuss \emph{how does the adaptation performance vary across different source-target combinations?}

First, the different adaptation performance can be traced back to the similarity between the source domain and the target domain. When the source and target datasets are alike, knowledge can be transferred to the target domain with less efforts. An example can be found in the LIAR $\rightarrow$ Constraint dataset in \Cref{tab:results}, where both datasets have similar input types and text lengths. In this case, the na{\"i}ve baseline achieves 0.7255 BA while \ours can reach up to 0.8147 in BA. In contrast, GettingReal $\rightarrow$ Constraint demonstrates a larger domain gap with different input types (News $\rightarrow$ Social Media) and length (738.9 $\rightarrow$ 32.7). Thus, we observe less improvements using \ours, with 0.7440 and 0.7416 in BA and accuracy respectively. The variations in the improvements can be further attributed the to the label shift between the source and target domain. When the source domain has a significantly different label distribution from the target domain, the bias may be carried over to the target domain during adaptation. An example can be found in GossipCop $\rightarrow$ CoAID, where the target label distribution is 10:90. As such, the baseline methods perform poorly in the balanced accuracy metric. Compared to the baseline methods, \ours achieves 0.6487 in BA in contrast to only circa 0.5 of baseline methods, suggesting the effectiveness of the proposed label correction component. Therefore, the domain discrepancy and the label shift between the source and target domains are observed to be crucial for the expected adaptation performance in COVID-19 misinformation detection.

\begin{table}[t]
\small
\centering
\begin{tabular}{@{}lccc@{}}
\toprule
\textbf{Dataset}                 & \textbf{\; BA $\uparrow$ \;} & \textbf{\; Acc. $\uparrow$ \;} & \textbf{\; F1 $\uparrow$ \;} \\ \midrule
\multicolumn{4}{c}{\textbf{CoAID}}                                                                                              \\ \midrule
\textbf{\ours}                   & \textbf{0.5932}              & \textbf{0.9160}                & \textbf{0.9554}              \\
\textbf{-Label Correction}       & \ul{0.5749}                  & \ul{0.9156}                    & \ul{0.9553}                  \\
\textbf{-Contrastive Adaptation} & 0.5513                       & 0.8676                         & 0.9251                       \\ \midrule
\multicolumn{4}{c}{\textbf{Constraint}}                                                                                         \\ \midrule
\textbf{\ours}                   & \textbf{0.7186}              & \textbf{0.7192}                & \ul{0.7334}                  \\
\textbf{-Label Correction}       & \ul{0.6745}                  & \ul{0.6819}                    & \textbf{0.7366}              \\
\textbf{-Contrastive Adaptation} & 0.5673                       & 0.5846                         & 0.7049                       \\ \midrule
\multicolumn{4}{c}{\textbf{ANTiVax}}                                                                                            \\ \midrule
\textbf{\ours}                   & \textbf{0.6652}              & \textbf{0.7104}                & \textbf{0.7870}              \\
\textbf{-Label Correction}       & \ul{0.6077}                  & \ul{0.6627}                    & \ul{0.7572}                  \\
\textbf{-Contrastive Adaptation} & 0.5281                       & 0.5899                         & 0.6890                       \\ \bottomrule
\end{tabular}
\caption{Ablation study of \ours.}
\label{tab:ablation}
\vspace{-7pt}
\end{table}

\subsection{Ablation Studies}
\textbf{Label Correction and Contrastive Adaptation}: We evaluate the effectiveness of the label correction component and contrastive adaptation by comparing our results from \ours to the results trained without label correction and contrastive adaptation. We 
report the performance on target datasets (averaged over all source datasets) in \Cref{tab:ablation}. Note that removing the contrastive adaptation component reduces \ours to the na{\"i}ve baseline, as the pseudo labeling and label correction are designed for the following contrastive adaptation stage.
As expected, we observe performance drops by removing the label correction component in \ours. On average, the performance reduces 6.1\% and 3.6\% in BA and accuracy when we remove the label correction component in \ours. Surprisingly, removing label correction slightly improves the F1 score on Constraint, 
this is because the model tends to predict true more frequently without label correction. As F1 does not considers true negatives, the value can be lower even if the model predicts with an increased fairness between both classes. Additionally, we observe that the majority of the improvements in CoAID are from the label correction component rather than the contrastive adaptation stage. A potential reason is the disproportionate label distribution in CoAID, where over 90\% data examples are credible information. Thus, the performance improves significantly by correcting the label shift between CoAID and the source datasets.

\begin{figure}[t]
\centering
\includegraphics[width=0.89\linewidth]{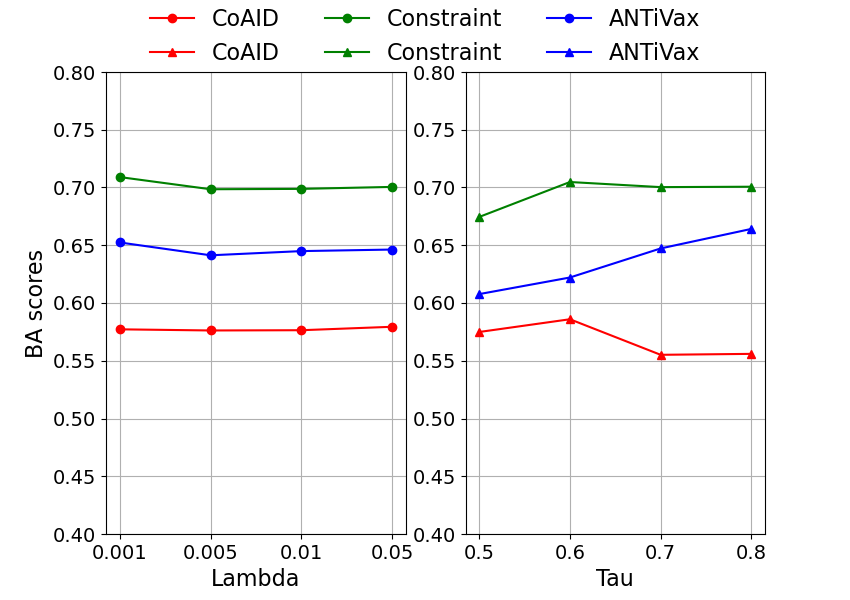}
\caption{Hyperparameter study for $\lambda$ and $\tau$.} 
\label{fig:hyperparameter}
\vspace{-7pt}
\end{figure}

\textbf{Sensitivity Analysis of Hyperparameters}: We study the sensitivity of the hyperparameters $\lambda$ and confidence threshold $\tau$. Similarly, we vary the input hyperparameter in adaptation and report the results on all target datasets. The reported numbers are BA values and are averaged across source datasets, see \Cref{fig:hyperparameter}. The results on hyperparameter $\lambda$ are comparatively stable and the BA scores are less sensitive to the changes of $\lambda$. For confidence threshold $\tau$ in pseudo labeling, the results vary across datasets. For example, the performance peaks for Constraint and CoAID at $\tau = 0.6$, while the results improve steadily with increasing $\tau$ for AnTiVax, suggesting further room for improvements on ANTiVax. Overall, the proposed \ours is robust to the changes of $\lambda$ and $\tau$ and consistently outperforms baselines even with sub-optimal hyperparameters.
\section{Conclusion}

In this paper, we propose a novel framework for domain adaptation in COVID-19 misinformation detection. We design \ours and propose to perform label shift correction and contrastive learning for early misinformation detection in the COVID domain. Unlike existing methods, we design a label correction component and adapt the conditional distributions to improve the target domain performance. Extensive experiments demonstrate that \ours is effective in domain adaptation for misinformation detection by achieving 11.5\% average improvements and up to 34.5\% improvements under label shifts between the source and target domains.


\section*{Acknowledgments}

This research is supported in part by the National Science Foundation under Grant No. IIS-2202481, CHE-2105005, IIS-2008228, CNS-1845639, CNS-1831669. The views and conclusions contained in this document are those of the authors and should not be interpreted as representing the official policies, either expressed or implied, of the U.S. Government. The U.S. Government is authorized to reproduce and distribute reprints for Government purposes notwithstanding any copyright notation here on.


\bibliographystyle{ACM-Reference-Format}
\bibliography{reference, anthology}










\end{document}